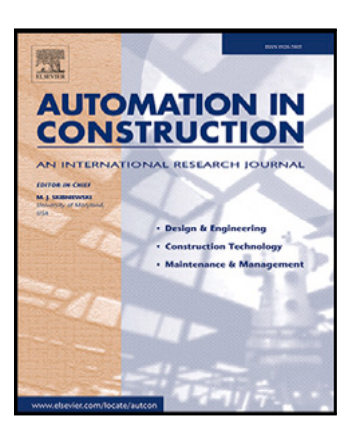

# Open-source automatic pipeline for efficient conversion of large-scale point clouds to IFC format

Slávek Zbirovský, Václav Nežerka *

*Faculty of Civil Engineering, Czech Technical University in Prague, Thákurova 7, 166 29 Praha 6, Czech Republic*

ABSTRACT

Building Information Model (BIM) creation usually relies on laborious manual transformation of the unstructured point cloud data provided by laser scans or photogrammetry. This paper presents Cloud2BIM, an open-source software tool designed to automate the conversion of point clouds into BIM models compliant with the Industry Foundation Classes (IFC) standard. Cloud2BIM integrates advanced algorithms for wall and slab segmentation, opening detection, and room zoning based on real wall surfaces, resulting in a comprehensive and fully automated workflow. Unlike existing tools, it avoids computationally- and calibration-intensive techniques such as RANSAC, supports non-orthogonal geometries, and provides unprecedented processing speed, achieving results up to seven times faster than fastest competing solutions. Systematic validation using benchmark datasets confirms that Cloud2BIM is an easy-to-use, efficient, and scalable solution for generating accurate BIM models, capable of converting extensive point cloud datasets for entire buildings into IFC format with minimal user input.

## 1. Introduction

The construction industry has undergone a rapid digital transformation over the past decade, ushering in a new era of architectural and engineering design. This transformation, commonly called Construction 4.0 [1], is driven by advancements in digital technologies that improve efficiency, quality, safety, and interoperability in construction projects [2]. A key aspect of this transformation is the increasing reliance on digital workflows to document and manage the lifecycle of buildings. However, these workflows remain challenging to implement in existing buildings due to the lack of reliable digital documentation.

The absence or obsolescence of building documentation presents a significant barrier to the efficient retrofitting and maintenance of existing structures. Accurately representing buildings in digital form is crucial for various applications, including energy efficiency upgrades, structural assessments, and facility management. Point cloud-based workflows are increasingly employed to capture detailed three-dimensional representations of buildings to address this. Laser scanning and photogrammetry are the primary technologies used for this purpose, generating high-density point clouds that record the spatial geometry of building elements with high precision [3,4].

Despite the high accuracy of point cloud data, raw point clouds lack semantic structure and are unsuitable for direct use in most engineering and architectural applications. Building information modeling (BIM) provides a structured digital representation of a building, integrating geometric and semantic information into a single framework. Converting point clouds into BIM, commonly referred to as scan-to-BIM, usually remains a labor-intensive task that relies heavily on manual input from skilled professionals [5].

Recent research efforts have sought to reduce the manual workload associated with scan-to-BIM by introducing methods providing various levels of automation. Semi-automatic approaches typically assist users in identifying structural elements but still require significant manual intervention [6]. Fully automated solutions aim to extract building components without user input, leveraging rule-based segmentation, deep learning, and geometric fitting techniques [7–11]. While these methods have demonstrated promising results, they often suffer from computational inefficiencies, limitations in handling complex geometries, or dependencies on specific commercial software environments.

The (semi-)automatic scan-to-BIM tools employ a combination of computational techniques to segment and classify building elements within point clouds. These include point cloud density histograms along selected axes [8,10,12,13], trajectory-based plane detection that relies on specialized mobile laser scanning systems [14], and deep learning models designed for structural element segmentation [15], though these primarily focus on classification rather than generating complete BIM-compatible geometries. Wall detection methods typically follow either a room-based or wall-based approach. Room-based methods identify enclosed spaces and approximate their boundaries [16,17], but often struggle with incomplete or missing data. In contrast, wall-based techniques reconstruct individual walls by detecting their surfaces and resolving spatial relationships [10,14,18–20], offering more precise 3D representations at the cost of additional preprocessing. A widely used strategy for extracting walls from point clouds involves the RANSAC algorithm, which detects planar surfaces [18,21–25], but requires careful parameter tuning and is sensitive to cluttered indoor environments. An alternative approach simplifies the scene into a 2D plan view, assuming vertical walls aligned to a regular grid [12,20,26,27], improving processing speed but limiting adaptability to irregular or non-standard building geometries.

* Corresponding author.
*E-mail address:* vaclav.nezerka@cvut.cz (V. Nežerka).

Based on the authors' experience, existing scan-to-BIM approaches face critical drawbacks that limit broader adoption. Many solutions exhibit high sensitivity to parameter settings, struggle in complex indoor environments, or require extensive pre-processing. Additionally, some methods impose strict assumptions about building layouts, such as strict orthogonality. Additionally, a significant portion of available tools depends on proprietary software ecosystems [11,28–30], restricting adaptability and limiting accessibility for a broader audience. While some projects have made their code publicly available [19], most remain tied to commercial platforms, reducing transparency and customization options. This reliance on closed ecosystems prevents researchers and practitioners from refining or adapting workflows to specific project needs, ultimately slowing innovation in the field. Furthermore, some seemingly promising approaches operate with significant constraints, such as reconstructing only visible surfaces rather than full volumetric elements [31].

This paper introduces Cloud2BIM, a fully open-source, Python-based scan-to-BIM tool that automates the conversion of large-scale point clouds into BIM models in the industry foundation classes (IFC) format, without relying on proprietary software. It features a fully automated processing pipeline that efficiently segments and classifies structural elements, including slabs, walls, and openings, while accurately handling non-orthogonal geometries. Unlike many existing methods, Cloud2BIM reconstructs building topology explicitly, ensuring that individual components are correctly structured. The direct IFC export enables seamless interoperability with BIM platforms such as Revit, Archicad, and FreeCAD, while also allowing compatibility with various free and open-source software tools for IFC editing, including Blender with the Bonsai plugin. Additionally, it integrates zoning capabilities, deriving room boundaries from actual wall surfaces for improved spatial accuracy in applications such as energy modeling or facility management. By offering a computationally efficient, scalable, and fully transparent workflow, Cloud2BIM removes barriers to adoption, allowing users to refine and adapt scan-to-BIM processes to diverse architectural and engineering needs.

## 2. Methods

This section outlines the algorithms used to obtain the geometric representation of building structures (Fig. 1) through Cloud2BIM, our Python-based software that adheres to open-source principles. A comprehensive list of the program dependencies and packages utilized during development is provided in Table 1. The entire codebase is accessible on our GitHub repository [32]. Fig. 2 presents a diagram illustrating the program workflow; the implementation of this workflow can be found in the *cloud2entities.py* file [32].

Cloud2BIM supports two types of file formats for input: .e57 and .xyz. These formats generally provide information about the position of individual points $(x, y, z)$. While the .E57 format is designed to include colors (r, g, b) and intensities as standard attributes, the .XYZ format may include colors and intensities, but this is not standardized and depends on the specific implementation. For accurate classification within Cloud2BIM, only the position data of the points are utilized.

**Table 1**
List of Python and associated packages used in the development of the Cloud2BIM software, including their versions and types of licenses.

| Software/Library | Version | License |
|---|---|---|
| Python | 3.10 | PSF license |
| Numpy | 1.26.2 | BSD license |
| Matplotlib | 3.8.1 | PSF license |
| Pye57 | 0.4.1 | MIT license |
| Pandas | 2.1.3 | BSD 3-Clause license |
| Open3D | 0.17.0 | MIT license |
| Tqdm | 4.66.1 | MIT license |
| Scikit-image | 0.22.0 | BSD 3-Clause license |
| Opencv-python | 4.8.1.78 | MIT license |
| Scipy | 1.11.3 | BSD 3-Clause license |
| Ifcopenshell | 0.7.0.231018 | GNU lesser general public license |

### 2.1. Testing datasets

Three distinct datasets were chosen to evaluate the proposed solutions and ensure an unbiased assessment of algorithm performance. The datasets comprised two own scans conducted by our team and one synthetic benchmark dataset. A Trimble X7 scanner was employed for the scans. A professional surveyor meticulously created these scans by establishing a geodetic network, stabilizing field points, determining coordinates with GNSS technology, and refining the network through terrestrial surveying. The resulting scans were of high quality and required no accuracy modifications, although they were thinned to reduce computational demands.

The process of data dilution was executed using the open-source CloudCompare software, where the point clouds were diluted to a prescribed minimum distance to the nearest point ($d_{\min}$). In 'spatial' mode, the user specifies a minimum distance between points. CloudCompare then selects points from the original cloud, ensuring no point in the resulting cloud is closer to another than the specified distance. As this value increases, fewer points are retained. The underlying algorithm operates as follows: Points are processed in an order determined by their spatial organization within an octree structure. The initial point is marked as 'to be kept'. For each subsequent point, the algorithm checks for neighbors within the specified radius. If no neighboring points are marked as 'to be kept', this point is also marked. Ultimately, only points marked as 'to be kept' are included in the final output cloud. The impact of this dilution is illustrated in Fig. 3.

Such an approach to dilution of point clouds proved effective for buildings with relatively simple geometric details. However, for buildings of historical significance that feature intricate decorations, the dilution strategy proposed by [33] is more suitable. This method adjusts the dilution factor based on the local curvature, helping to preserve critical features while reducing the risk of feature loss, often referred to as 'feature washing.' Caution is advised with this approach as it tends to remove points from planar surfaces, which could lead to insufficient density for effective point cloud segmentation. Although avoiding dilution prevents data loss, it dramatically increases computational demands and may not always be practical.

### 2.2. ETH Zurich synth3

The synthetic dataset, provided by the Multimedia Laboratory at the University of Zurich [34], serves as a challenging benchmark for volumetric wall reconstruction. It is designed to evaluate methods that operate under the Atlanta-world assumption, which allows for non-orthogonal wall configurations (Table 2). This assumption permits the depiction of scenes using vertical and horizontal planes in 3D, without the constraint of vertical planes (walls) being orthogonal to each other [35] as assumed, e.g., by Kong et al. [36] or Kim and Lee [37] in their recent studies on classification of point clouds. The dataset encompasses 20 interior walls, features 13 door openings, and is designed for algorithms that reconstruct interior walls connecting at various angles. Notably, the external surfaces of the object were not created by the dataset authors, simulating purely indoor scanning. This dataset has been pivotal in research focused on indoor wall detection, contributing significantly to other studies [18,19,38].

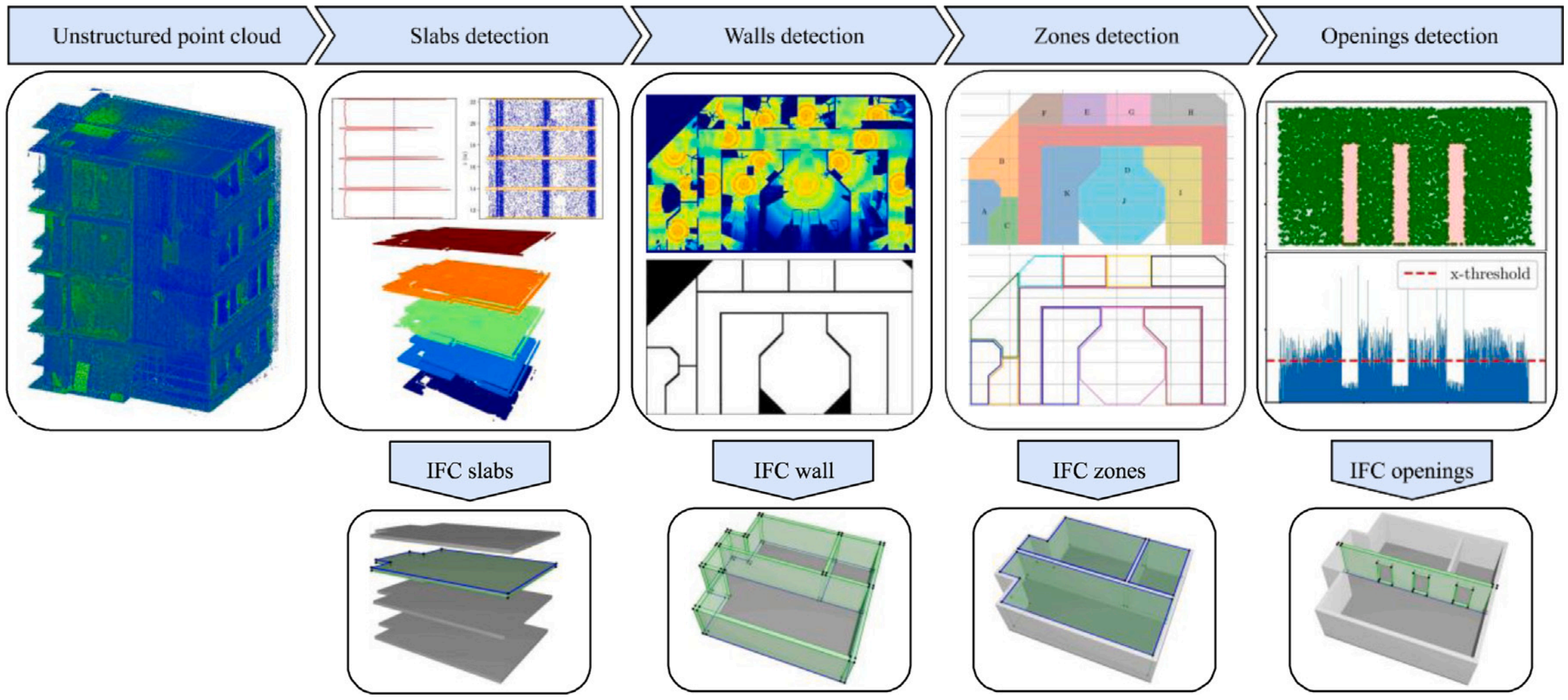

**Fig. 1.** Overview of the Cloud2BIM workflow, illustrating the automated conversion from raw point clouds to IFC BIM models, including detection of slabs, walls, zones, and openings, and their conversion to the IFC format.

**Table 2**
ETH Zurich synth3 dataset specification.

| Dataset name | Visualization | $n_{points}$ | Size | $d_{min}$ |
|---|---|---|---|---|
| ETH Zurich synth3 | | 6,905,684 | 468 kB | 0.01 |

**Table 3**
Kladno station dataset specification.

| Dataset name | Visualization | $n_{points}$ | Size | $d_{min}$ |
|---|---|---|---|---|
| Kladno station | | 2,551,721 | 190 kB | 0.025 |

### 2.3. Kladno station

Point cloud data from Kladno railway station (Table 3), captured using the terrestrial laser scanner, was employed to demonstrate the wall segmentation capabilities of the developed software on a real-world dataset. For this purpose, one floor of the building was utilized. Using an actual point cloud from this building provides the benefit of including real-world irregularities and imperfections that the program must navigate, thereby enhancing the robustness and applicability of the segmentation results. The complete original point cloud is available at Zenodo platform [39].

### 2.4. Prague hotel

The point cloud dataset representing a former hotel in Prague (Table 4) encompasses an eleven-story building section. This dataset was chosen to demonstrate the software's ability to manage large datasets and to highlight the precision and efficiency improvements afforded by density-based height segmentation. The laser scanning was conducted from inside the building using a terrestrial laser scanner, resulting in the building envelope being largely absent, except for areas visible through window openings.

### 2.5. Identification of slabs

Cloud2BIM adopts a widely used method that analyzes point density within a histogram to identify horizontal planes in point clouds [19], illustrated in Fig. 4. This approach assumes that the floor slab is horizontal relative to the $x$-$y$ plane, has uniform thickness, and lacks local height variations like steps. A candidate for a horizontal surface is identified within a height range that contains at least 50% of the maximum number of points found in any single plane. This threshold is adjustable to account for uneven point distribution often seen in real-world scans, where fewer points may be recorded at floor level due to obstructions like furniture or scanner positioning. Similar approach was used e.g. by Chen et al. [40], who estimated stiffness of structures based on point cloud density distribution in order to assess their performance during earthquakes.

The floor-plan shape of the slab is then determined based on the surface index:

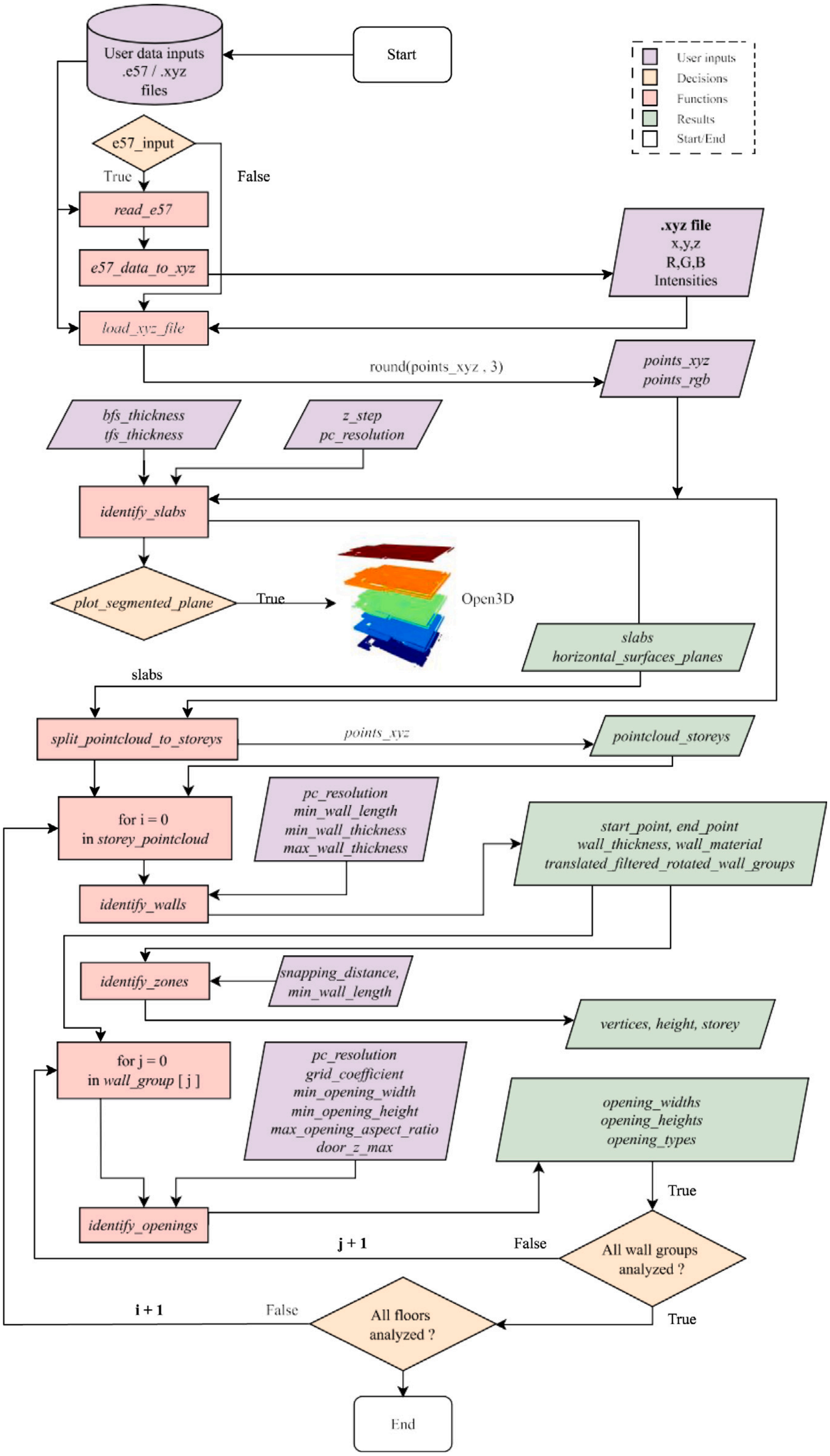

**Fig. 2.** Process of generating a 3D model of a building from laser scans, detailing each step from data acquisition to model completion.

1. The first slab, typically representing either a ground-level or a partially scanned slab, is assumed to have only its upper surface and requires a manual thickness setting.
2. If an even number of surfaces remains, it implies that both the ceiling slabs and possibly the roof have been scanned from both surfaces. Here, the surfaces are merged to create a floor plan point cloud for robust analysis, assuming that the slab mirrors the ceiling shape of the floor below (Fig. 5).
3. An odd number of surfaces suggests that the last surface is unpaired, potentially indicating an unscanned or non-flat roof, which also requires a manual thickness setting.

To process the point cloud, the function *create_hull_from_histogram* projects the 3D points onto a 2D plane, discarding the $z$-coordinate, and creates a 2D histogram (Fig. 6) that assists in forming a binary mask. Morphological operations — dilation and erosion — are then applied to refine the mask contours, maintaining object shape while smoothing irregularities. The mask is adjusted to ensure precise contour alignment with the grid.

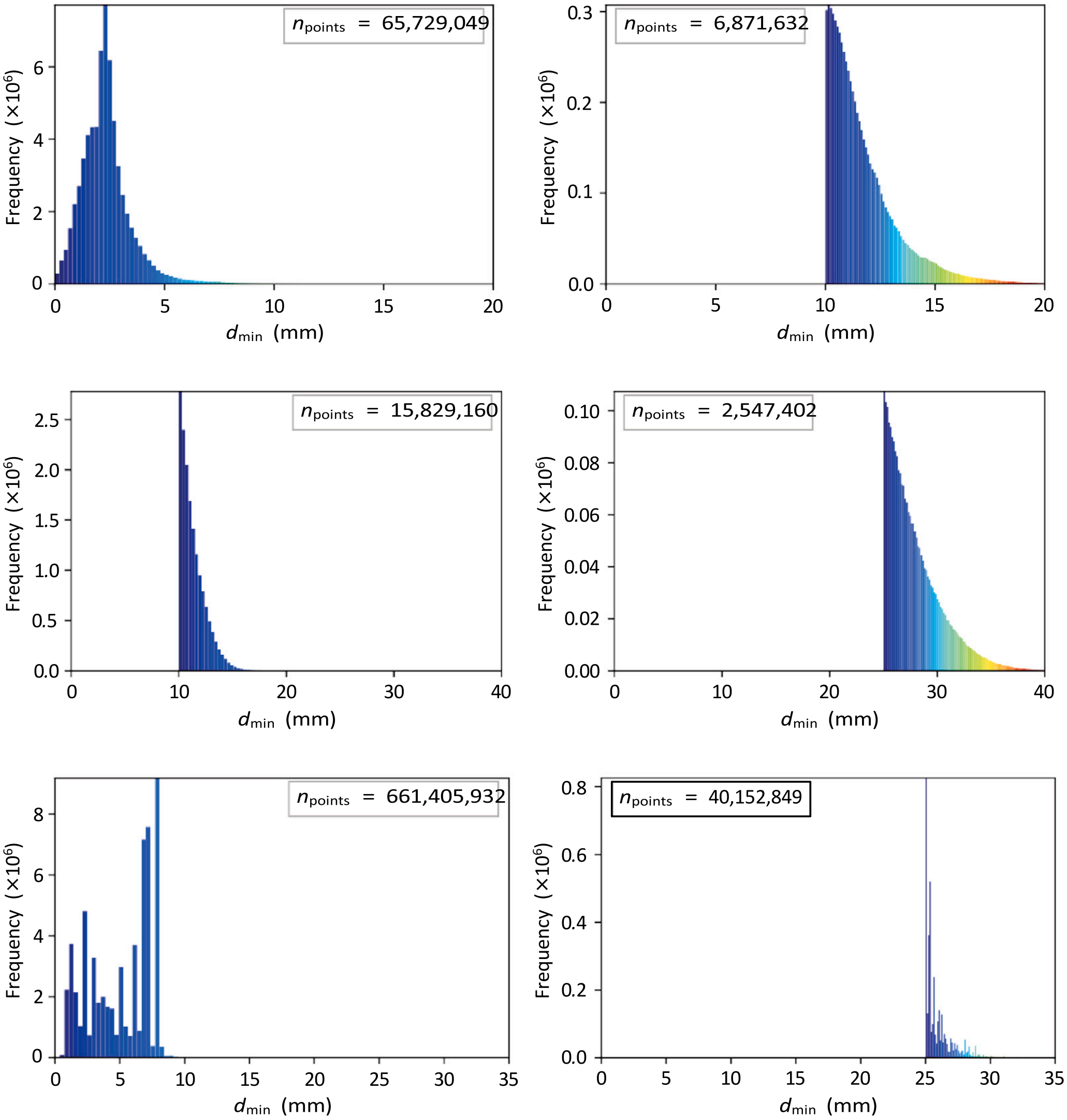

**Fig. 3.** Histogram of minimum distances to the nearest neighbor before (left) and after (right) dilution; ETH Zurich synth3 dataset (top), Kladno station dataset (middle), and Prague hotel dataset (bottom).

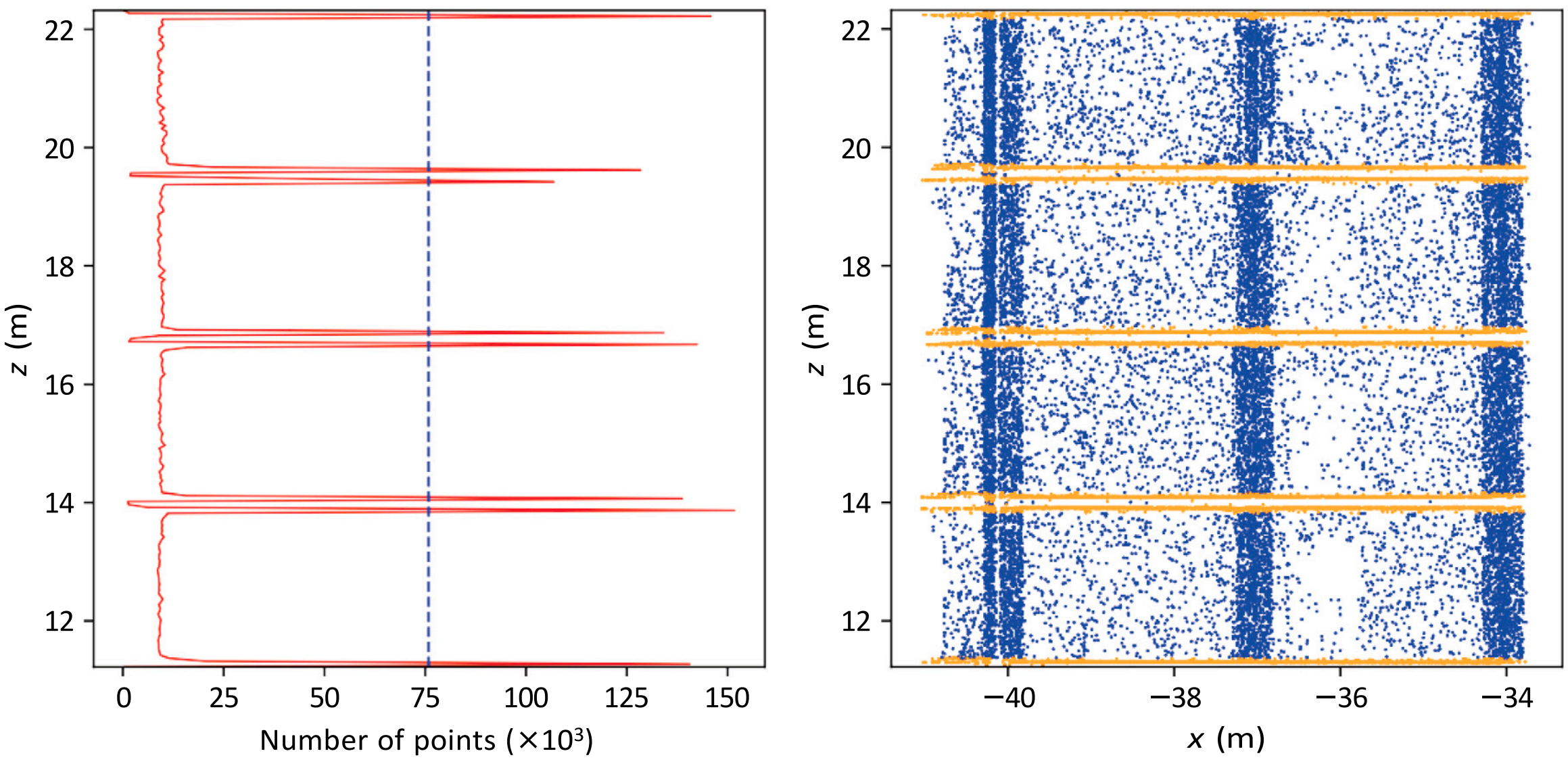

**Fig. 4.** Point density histogram (left) showing point counts by height ($z$-coordinate) with a 50% maximum threshold marked by a dashed blue line. The side view of the point cloud (right) displays point distribution in the $x$-$y$ plane, identifying horizontal surface candidates (orange) above the threshold and other points (blue) below it.

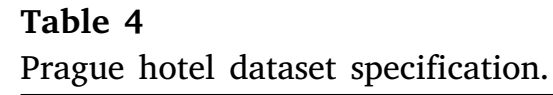

**Table 4**
Prague hotel dataset specification.

| Dataset name | Visualization | $n_{\text{points}}$ | Size | $d_{\min}$ |
|---|---|---|---|---|
| Prague hotel | | 40,152,849 | 2.54 GB | 0.025 |

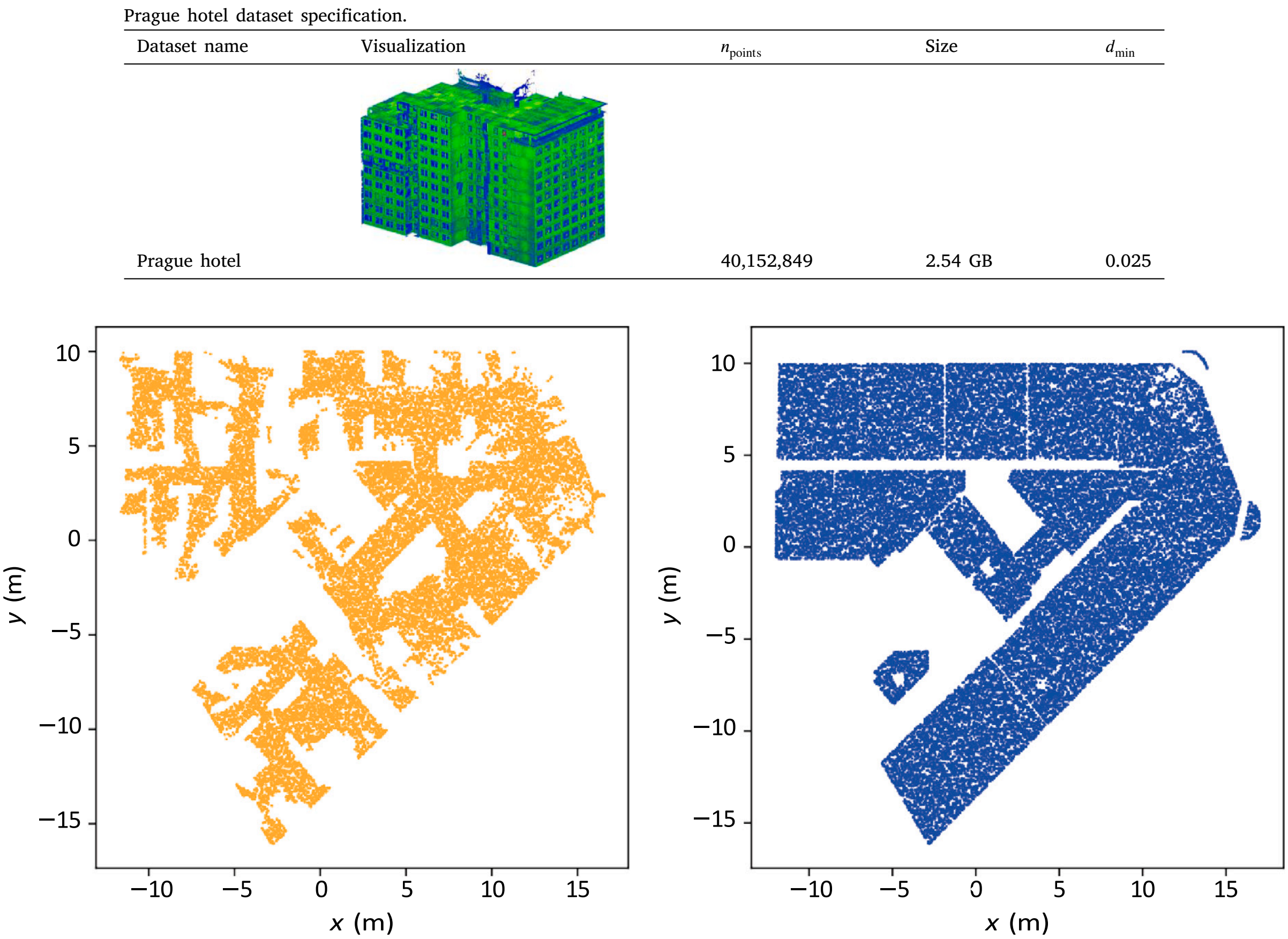

**Fig. 5.** Candidate point cloud subsets for merging representing the floor surface (left) with lower density due to obstructions during scanning and ceiling surface (right) used to create a floor plan when combined.

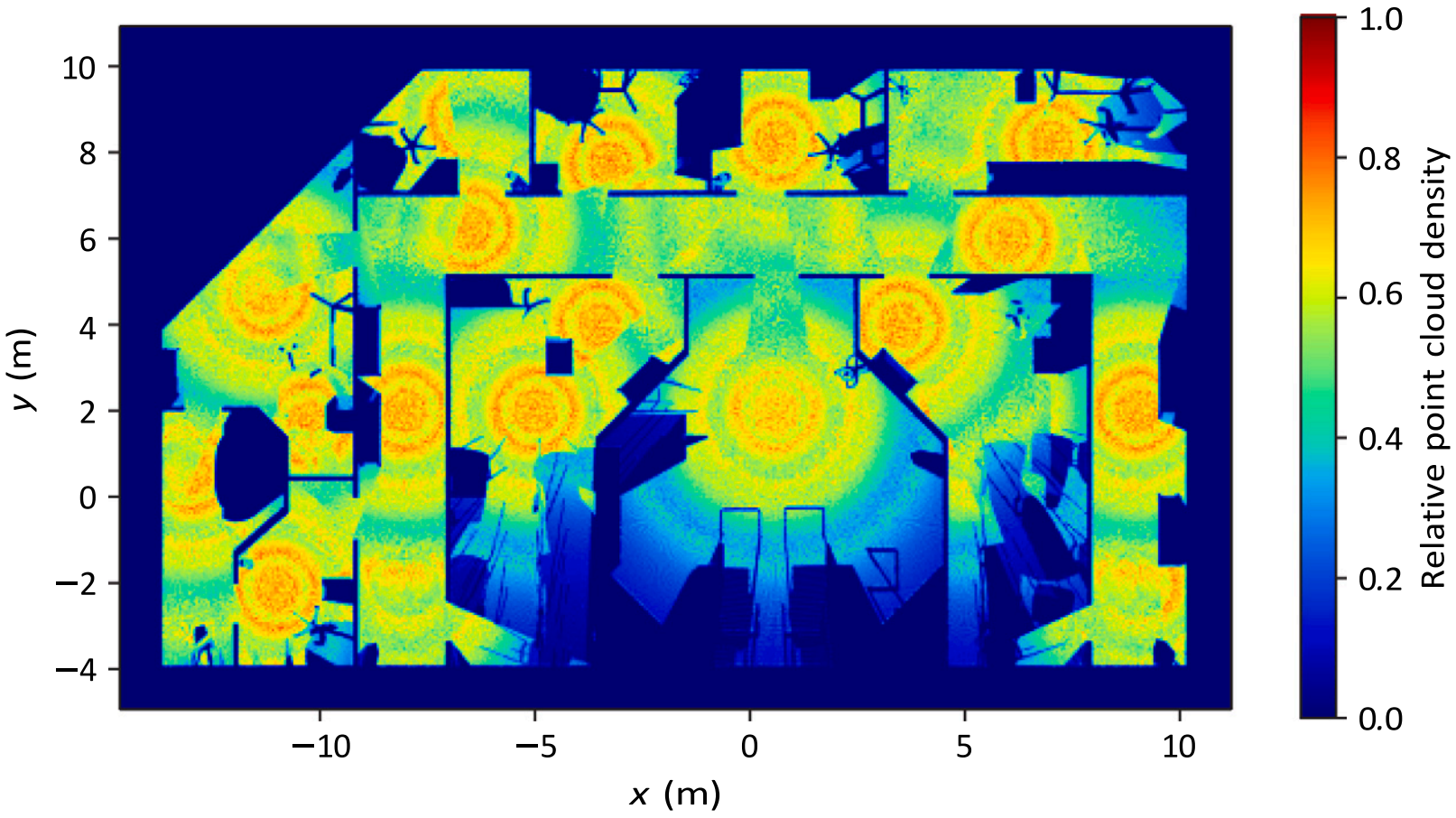

**Fig. 6.** 2D histogram of cloud point density in the identified floor plane (ETH Zurich synth3 dataset).

Using the algorithm by Suzuki and Abe [41] implemented through OpenCV's *findContours()* function, the most prominent contour is extracted, scaled, and converted into a Polygon object that delineates the floor plan. To simplify this contour, the Douglas–Peucker algorithm [42,43] is applied, reducing the number of points and smoothing the shape, as shown in Fig. 7. The algorithm iteratively reduces the number of points defining a polyline and operates by recursively eliminating points that deviate less than a user-defined tolerance, epsilon, from the simplified line segments. This process effectively represents the polygon of a slab with fewer, more significant points. The complete program workflow for slab segmentation is depicted in Fig. 8 and the implementation of this workflow can be found in the *aux_functions.py* file, function *indentify_slabs*, lines 196–307 [32].

### 2.6. Segmentation into storeys

Once the horizontal structures are identified, the point cloud is segmented into individual storeys. This segmentation facilitates the spatial organization of the building. The segmentation process is handled by the function *split_pointcloud_to_storeys*, which requires two primary inputs: *points_xyz* for the 3D coordinates of the point cloud, and *slabs* (Fig. 9). The implementation of this workflow can be found in the *aux_functions.py* file, function *split_pointcloud_to_storeys*, lines 310–328 [32]. Each slab is represented by a dictionary detailing the polygon of the slab, its $x$- and $y$-coordinates, its lower $z$-coordinate, and its thickness. The function processes each slab sequentially, extracting points between the base of one slab and the top of the slab below it while maintaining by a default setting 10 cm clearance from both the ceiling and the floor. The output is a list of point cloud subsets, where each subset corresponds to a distinct storey within the building.

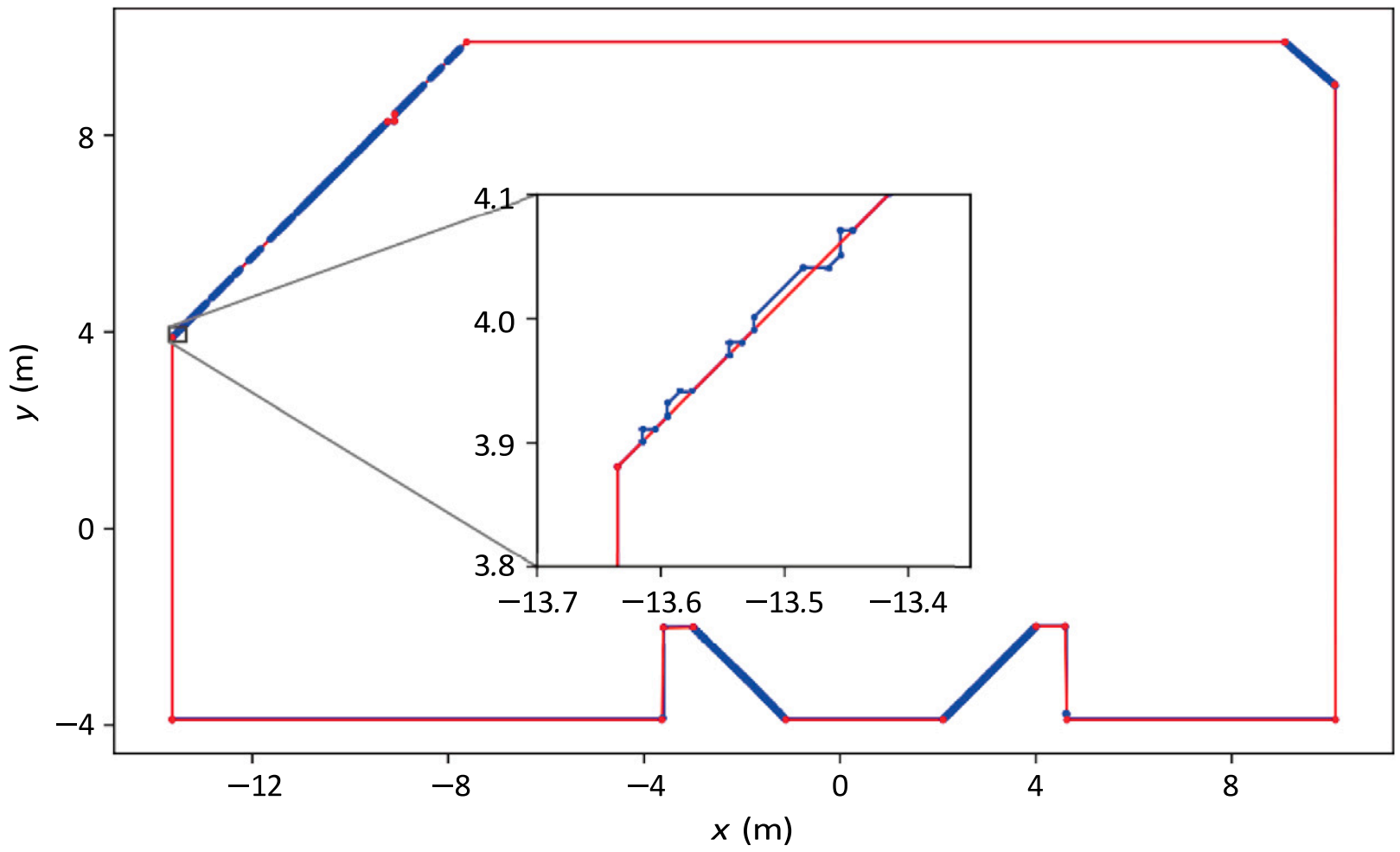

**Fig. 7.** Close-up view of the ceiling slab edge, with the original edge indicated in blue and the smoothed edge, generated using the Douglas Peucker algorithm in red.

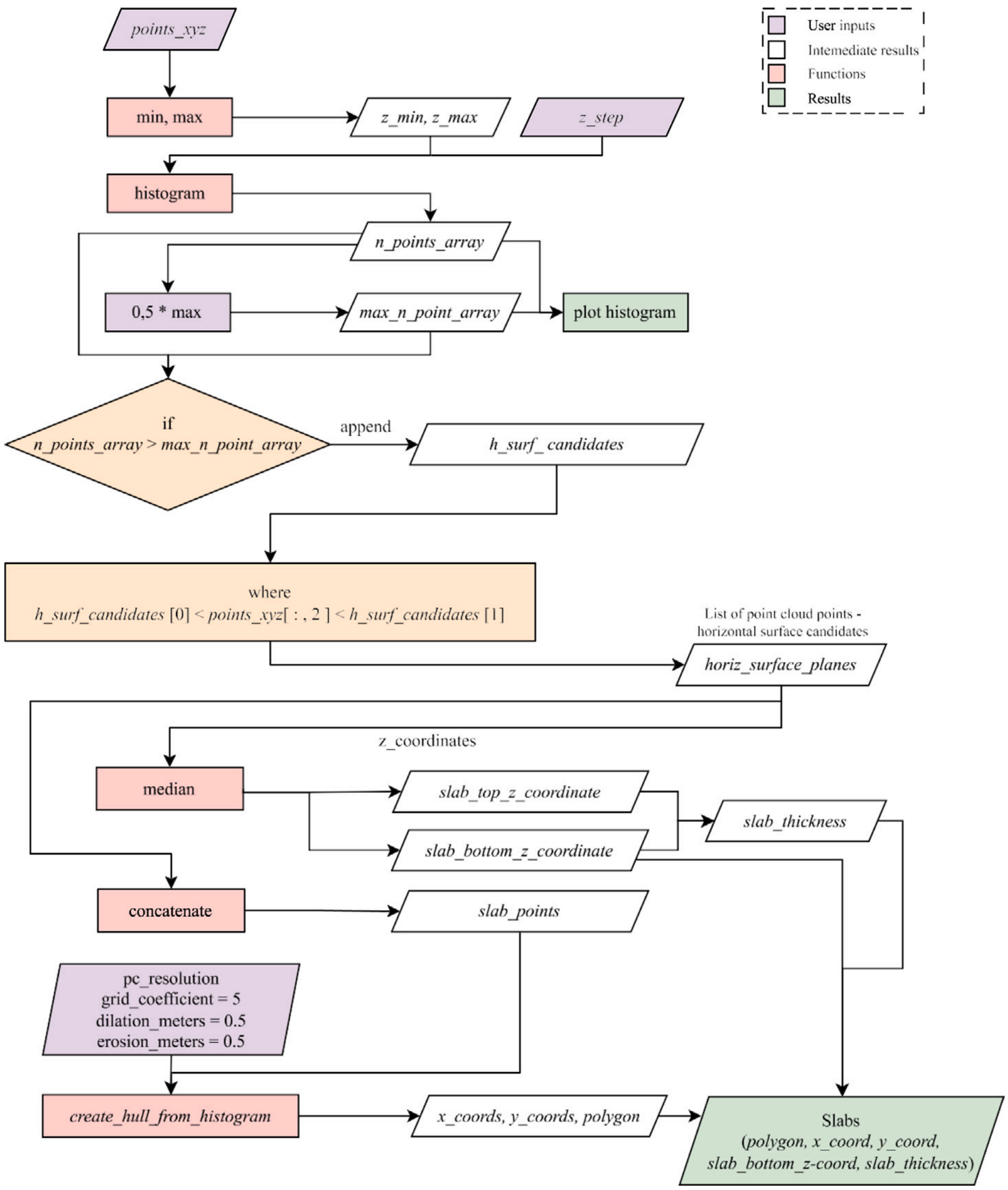

**Fig. 8.** Workflow of the function used to detect slabs in an unorganized point cloud.

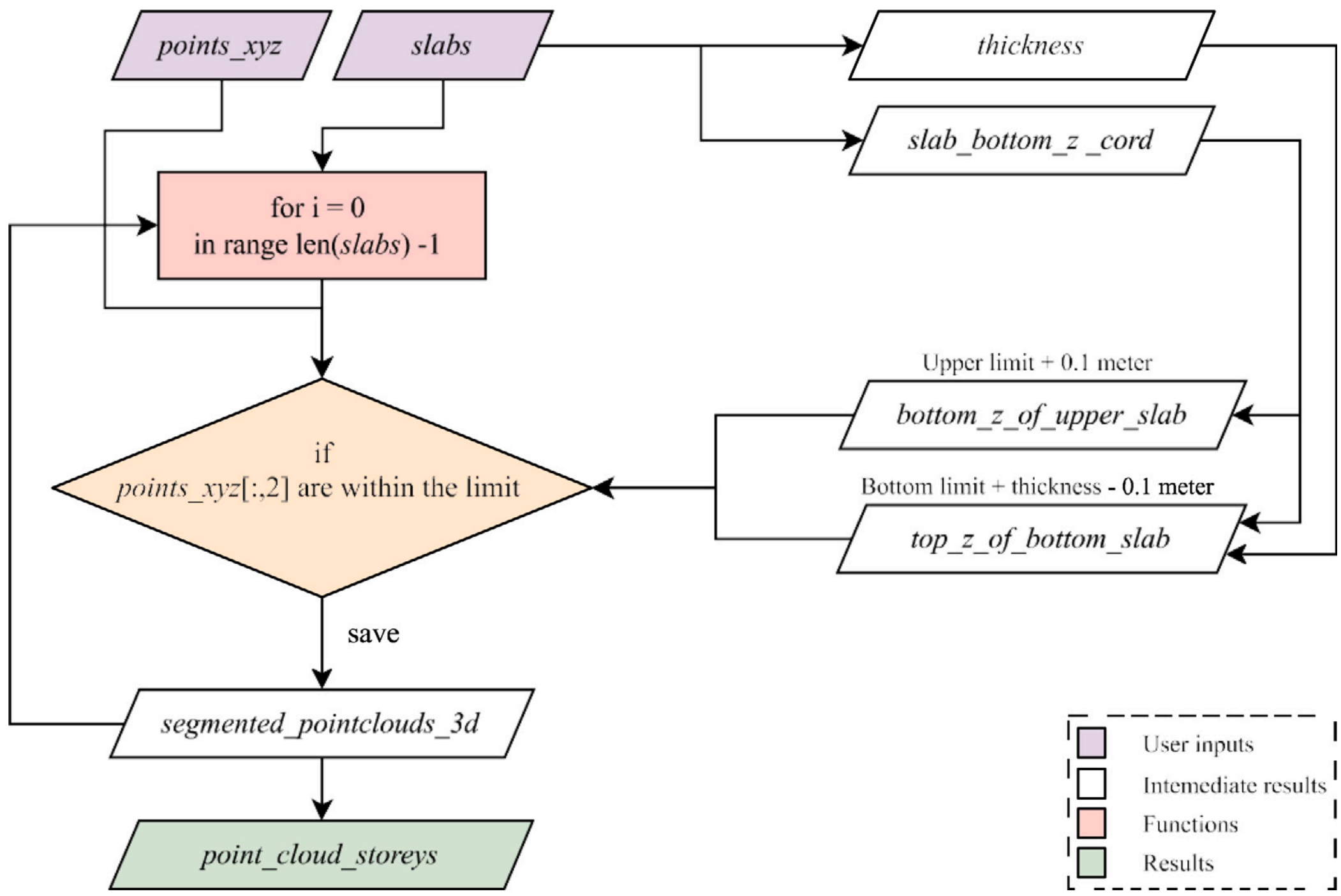

**Fig. 9.** Process of segmenting a point cloud into individual storeys, highlighting the sequence of operations from slab identification to storey-specific point cloud extraction.

### 2.7. Identification of walls

Two primary strategies prevail for reconstructing interior walls from point cloud data: room-based and wall-based. The room-based approach simplifies the segmentation process by focusing on reconstructing the geometry of individual rooms, which involves segmenting the point cloud into distinct spaces and determining the optimal shape for each. This approach, however, struggles with flexible parametric wall geometry and unenclosed spaces. On the other hand, the wall-based approach clusters point cloud data around each wall, aiming to reconstruct topologically consistent structures [19].

The Cloud2BIM's *Identify_walls* function, described using the flowchart in Fig. 10, detects and defines wall geometry within a point cloud. The implementation of this workflow can be found in the *aux_functions.py* file, function *identify_walls*, lines 815–970 [32]. It isolates individual wall segments within each floor's horizontal section, set by default to 90%–100% of the floor height to minimize noise and disruptive objects. The function begins by filtering points that fall within these height limits and proceeds to extract corresponding $x$- and $y$-coordinates, forming a 2D array (*points_2d*) that represents the horizontal section of the point cloud at the specified height range.

This 2D point cloud then undergoes histogram analysis to create a 2D histogram, where each bin's value reflects the point density in that spatial region. Converting this histogram into a binary image based on a predefined threshold isolates regions with high point densities (likely indicative of walls). This binary image (Fig. 11) serves as the basis for further wall detection and geometry definition. It undergoes processing to extract contours using the *findContours* function from the OpenCV library [44], explicitly targeting the outermost boundaries of wall segments.

As for slabs, each detected contour is approximated into simpler line segments using the Douglas–Peucker algorithm. The choice of epsilon significantly affects the quality of the resulting geometry: a very small epsilon value results in contours with many short segments, leading to an increase in computational time during the subsequent pairing of opposing wall surfaces to form volumetric walls (although these short, colinear segments are merged afterward). Conversely, an excessively large epsilon value may erroneously merge segments that represent separate surfaces, such as distinct walls near openings or shorter wall segments. Therefore, to preserve the geometric accuracy, epsilon must be carefully selected and should never exceed the length of the smallest segment intended to be represented. Our analysis indicates a reasonably wide range of epsilon values within which optimal contour simplification can be reliably achieved. Fig. 12 demonstrates that small epsilon values produce detailed and ragged contours with an excessive number of segments, while large epsilon values overly simplify contours, potentially merging separate surfaces near openings and short segments

Following segmentation, the wall segments undergo further analysis to join into longer segments based on their mutual angles, distances, and alignment. To form a volumetric representation, opposite wall surfaces are paired if they are parallel, within a maximum distance, and have sufficient overlap. A virtual surface is generated to complete the pairing if an outer surface is missing.

The final steps involve calculating the wall axis and adjusting connections to ensure correct topology. The algorithm identifies the longest segment as a reference, calculates its direction vector, and uses it to estimate the central position of the wall axis. Endpoints of wall segments are adjusted for intersections within the threshold of half the maximum wall thickness to ensure walls connect seamlessly.

The function outputs each wall's start and end points along with its thickness. The wall height is assumed to be the full height of the floor, as this method does not allow for the detection of shorter walls.

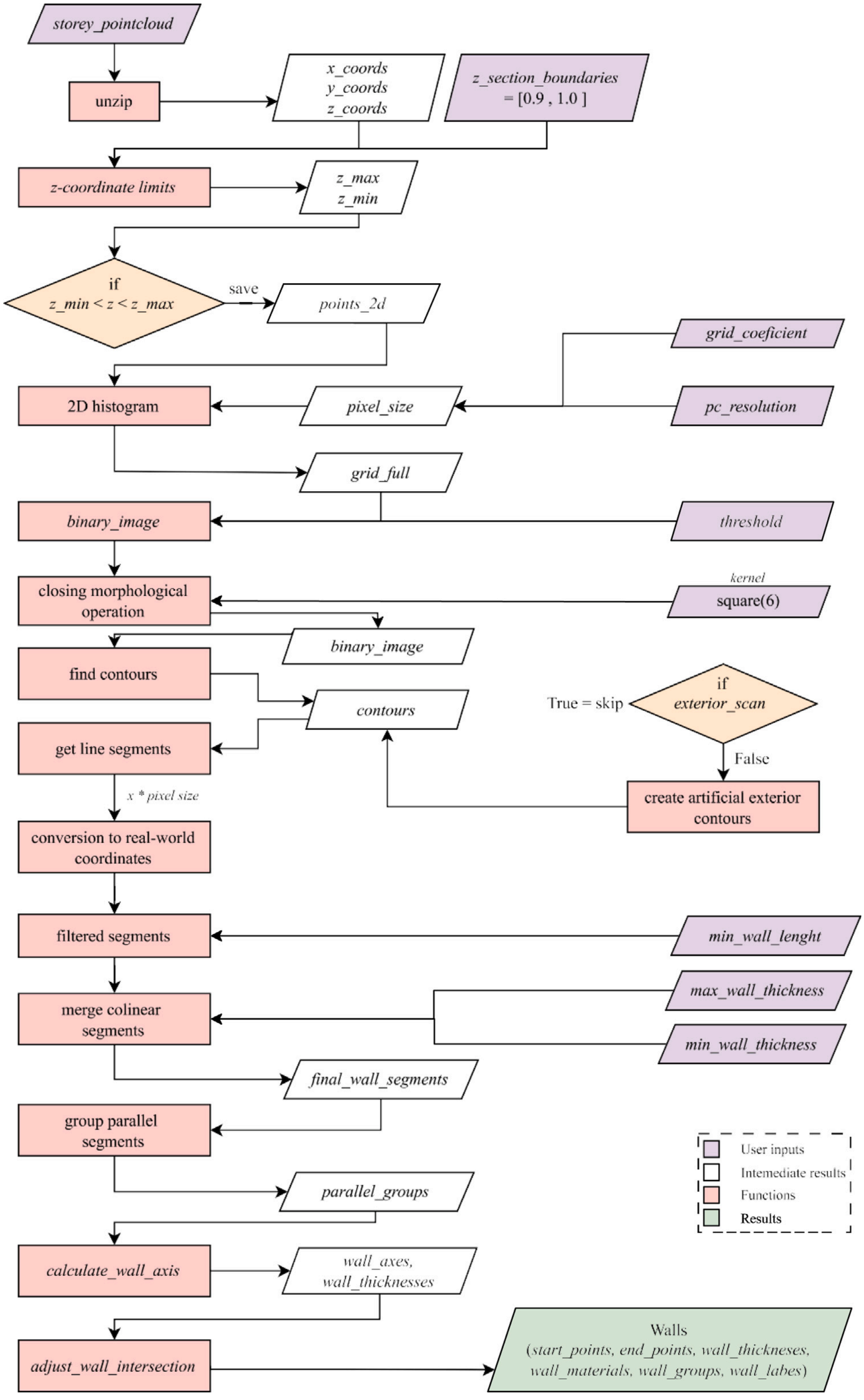

**Fig. 10.** Algorithms for segmentation of walls.

### 2.8. *Identification of zones*

Zones are entities related to the internal areas of individual rooms, typically bounded by floor slabs, ceilings, and surrounding walls. Zones facilitate the extraction of area and volumetric parameters for individual rooms and enable the assignment of various attributes related to their function, cladding materials, fire-safety parameters, acoustics, and other relevant aspects.

The detection of zones is typically approached through three primary methods [45]: (i) slice-based modeling, (ii) mesh simplification modeling, and (iii) floorplan-based modeling. Once the exact positions and geometries of walls are identified, Cloud2BIM circumvents these computationally intensive methods. Instead, the *identify_zones* function (Fig. 13) analyzes wall geometry by first processing the axes of walls to ensure connectivity and proper alignment. The implementation of this workflow can be found in the *space_generator.py* file, function *identify_zones*, lines 887–928 [32]. Using this function, wall segments are split at intersections of these axes and extended to connect with adjacent axes within a distance specified by the *snapping_distance* parameter. Subsequently, segments parallel to the wall axes, representing the surfaces of the walls, are generated at a distance equal to half the width of the individual walls. These parallel segments are then extended to intersect fully, forming the boundaries of zones. The function ultimately creates closed polygons defined by the wall axes; each polygon serves as a candidate zone. The complete process of zone identification is illustrated in Fig. 14 using the ETH Zurich synth3 (left) and Kladno station (right) datasets.

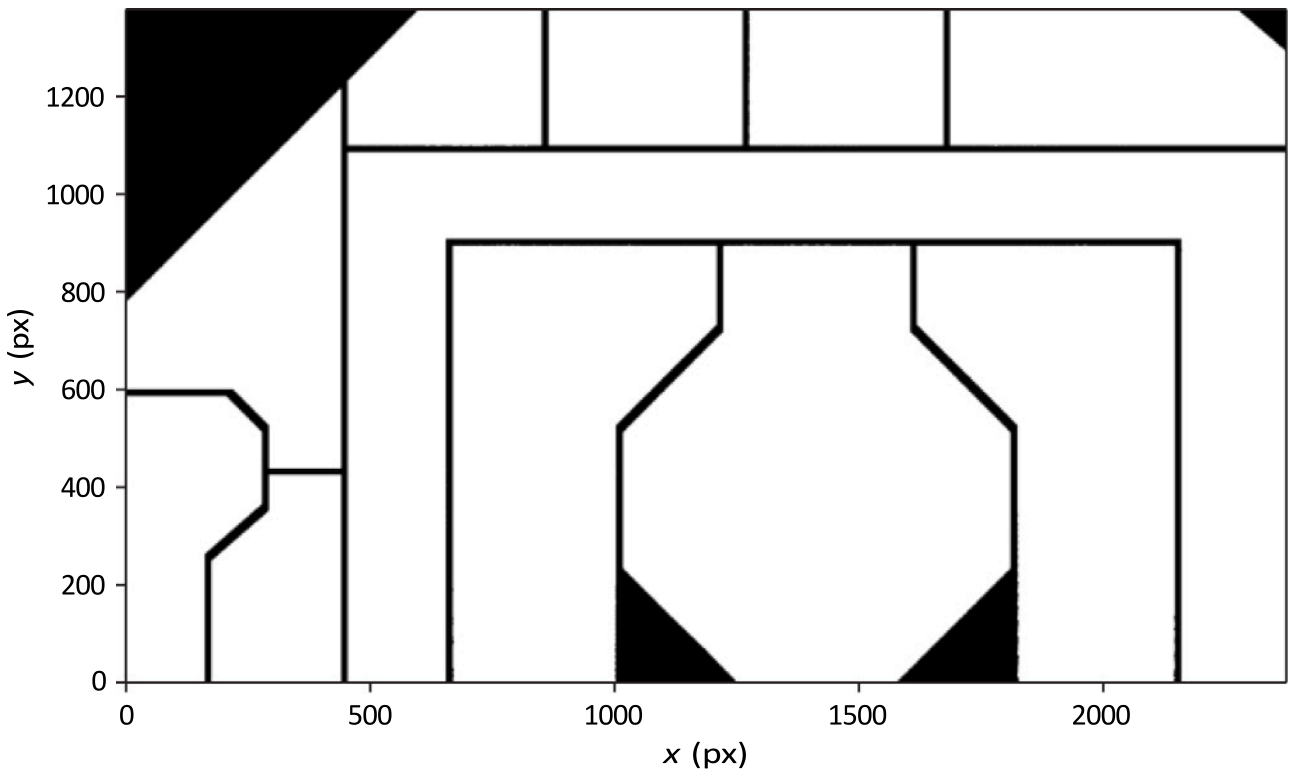

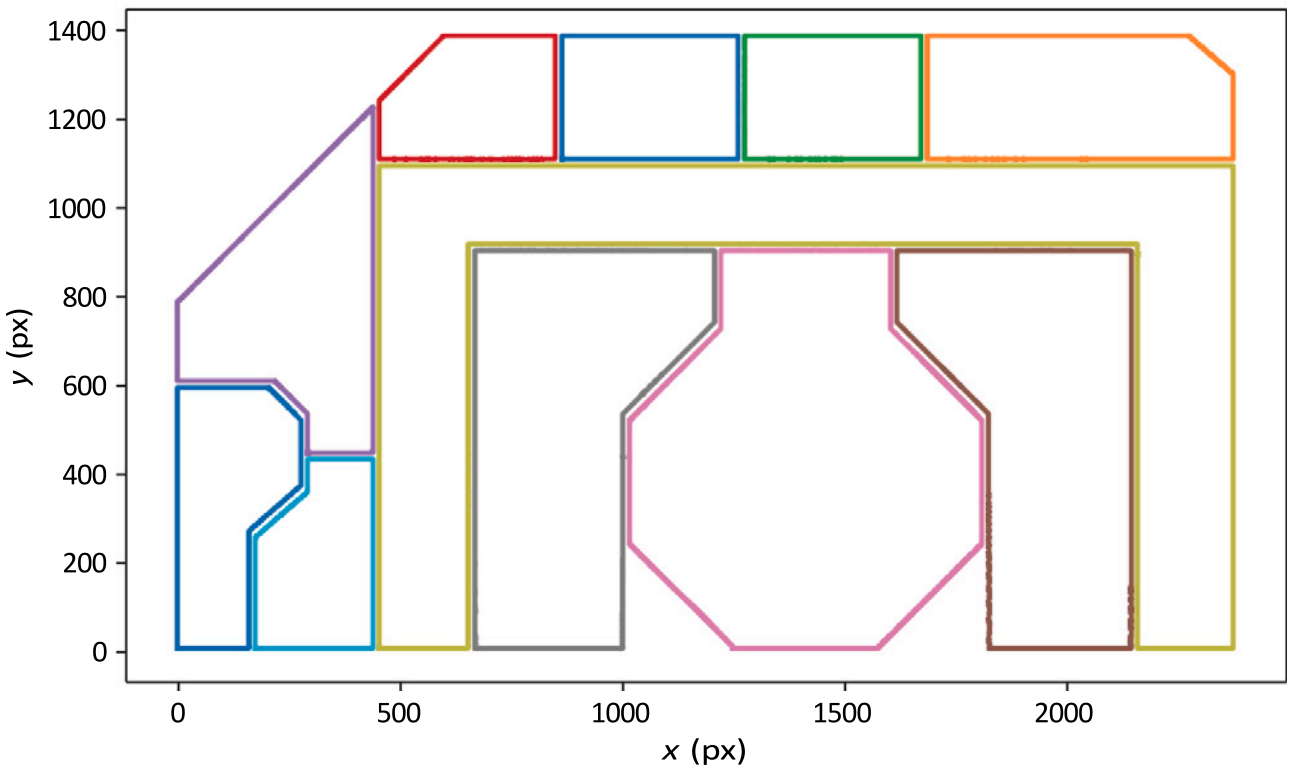

**Fig. 11.** Binary mask highlighting high-density regions indicative of wall positions, derived from a 2D histogram of the point cloud (top) and contours delineating wall segments in the 2D floor plan, derived from the binary mask segmentation (bottom); ETH Zurich synth3 dataset.

### 2.9. Identification of openings

Several methods for detecting openings have been developed over the years. Techniques for identifying door openings have typically been categorized into histogram-based approaches [11,28] and those that utilize scanner trajectory data [14]. Each method, however, comes with specific limitations. Density-based analysis can be sensitive to obstructions like furniture, affecting the histogram's accuracy, while trajectory-based methods depend heavily on particular scanning technology. Due to these constraints, our approach utilizes a density analysis from histograms, as illustrated in the flowchart in Fig. 15, where the function *identify_openings* is outlined. The implementation of this function can be found in the *aux_functions.py* file, function *identify_openings*, lines 1181–1331 [32].

Several heuristic rules enhance the histogram-based method in Cloud2BIM to pinpoint door and window openings accurately. These rules consider the sill position to differentiate between windows and doors, the aspect ratio of the opening, and the maximum and minimum widths of the opening. However, this approach restricts the detected shapes to rectangular openings.

The analysis begins by positioning the wall at the coordinate origin and rotating it into the $x$-$z$ plane. The floor plan position of the opening, primarily along the $x$-axis, is first determined. The histogram is then analyzed based on a threshold defined as a percentage of the tenth maximum value recorded to identify the start and end of gaps or low-density areas. The tenth maximum was selected based on observations of the histograms and found appropriate to avoid extreme maxima that seldom occur in the histogram; in some walls there were about 1–3 extreme maxima so that the tenth maximum provides a safe reserve, but the number of selected maximum can be changed. When matching openings are detected, this procedure is repeated for the $z$-axis histogram. Fig. 16 presents a side view of the wall, displaying histograms for the $x$ and $z$ axes with marked door openings in red. The effectiveness of this analysis is improved by the prior dilution of the point cloud to the minimum distance, as discussed in Section 2.1. A more uniform point distribution minimizes the risk of misjudging the thresholds for empty space in the histograms, thereby improving the accuracy of opening detection.

### 2.10. Creation of IFC output file

The industry foundation classes (IFC) file format is primarily designed to facilitate data transfer among various participants in the modeling process, allowing the sharing of the building information model and its associated data. Utilizing IFC as the final format to transfer data from point clouds to BIM is a well-established approach, as evidenced by similar implementations by Martens and Blankenbach [10], which included the reconstruction of slabs, floors, and walls. There are numerous scan-to-BIM methodologies discussed in the literature, such as those by Adán et al. [46] and Truong-Hong and Lindenbergh [47]; however, these approaches typically do not support direct export to IFC or any standardized BIM format. The capability to export directly to IFC, as implemented in our Cloud2BIM tool, facilitates seamless integration with a wide range of BIM software platforms. This compatibility significantly enhances the utility of the generated models for architectural planning, structural analysis, and lifecycle management, ensuring that stakeholders can leverage the full spectrum of BIM functionalities for efficient project execution and maintenance.

Our output of Cloud2BIM is generated in the IFC 4 ADD2 TC1 format using the IfcOpenShell library for Python [48]. This process includes specifying all necessary attributes for each IFC class. We chose the DesignTransferView_V1.0 Model View Definition to ensure that our models are compatible with and can be further edited in BIM software platforms such as ArchiCAD [49], Revit [50], or Allplan [51].

The structural hierarchy of IFC elements mirrors standard project organization practices as each IFC file must define precisely one instance of *IfcProject* or another *IfcContext*, such as *IfcProjectLibrary*, which serves as the anchor for all other data within the file. *IfcProject* includes essential attributes such as unit definitions, project name, and project phase, while other IFC classes like *IfcSite*, *IfcBuilding*, and *IfcBuildingStorey* help in assigning objects to their precise locations. In terms of geometry, the *IfcSlab* class is used to represent ceiling slabs, where the slab's perimeter is defined by an *IfcArbitraryClosedProfileDef*, typically a polyline and extruded along a specified axis using the *IfcExtrudedAreaSolid* entity. Walls are similarly modeled using the *IfcWall* class, with geometry defined by start and end points via the *IfcCartesianPoint* class, and extruded using the *IfcExtrudedAreaSolid*. Openings in the walls are handled by the *IfcOpening* class, linked to the parent wall through the *IfcRelVoidsElement* relationship, with their positions defined by the *IfcAxis2Placement3D* and modeled using the swept solid technique. Zones are created using the same techniques as slabs using the *Ifcspace class*. They are based on *IfcArbitraryClosedProfileDef*, which defines the area of the zone, and *IfcExtrudedAreaSolid*, which defines the height of the zone extrusion. Graphical representation of these classes is provided in Fig. 17.

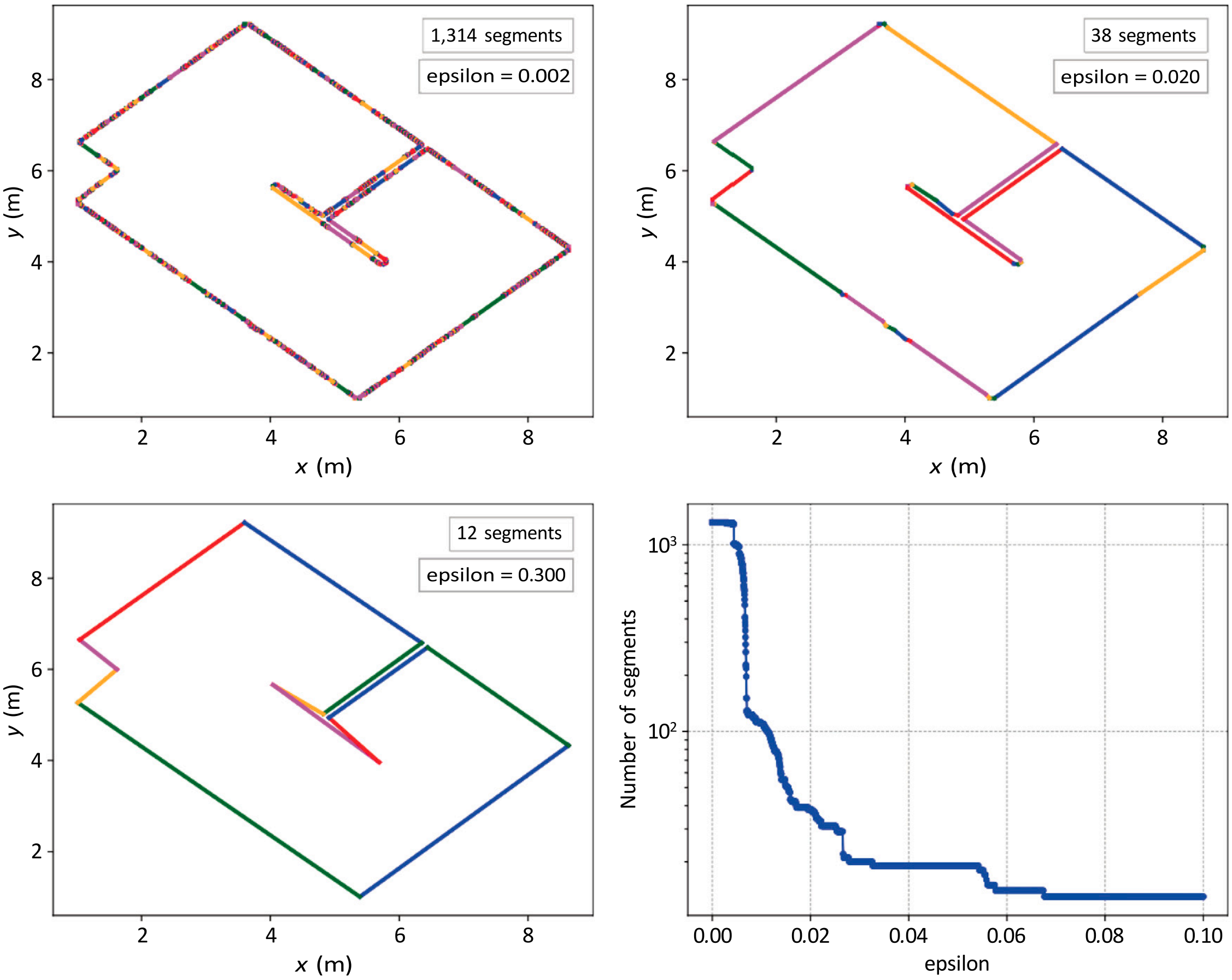

**Fig. 12.** Douglas–Peucker algorithm applied to contour approximation, demonstrating the influence of the epsilon parameter on contour simplification and wall surface identification.

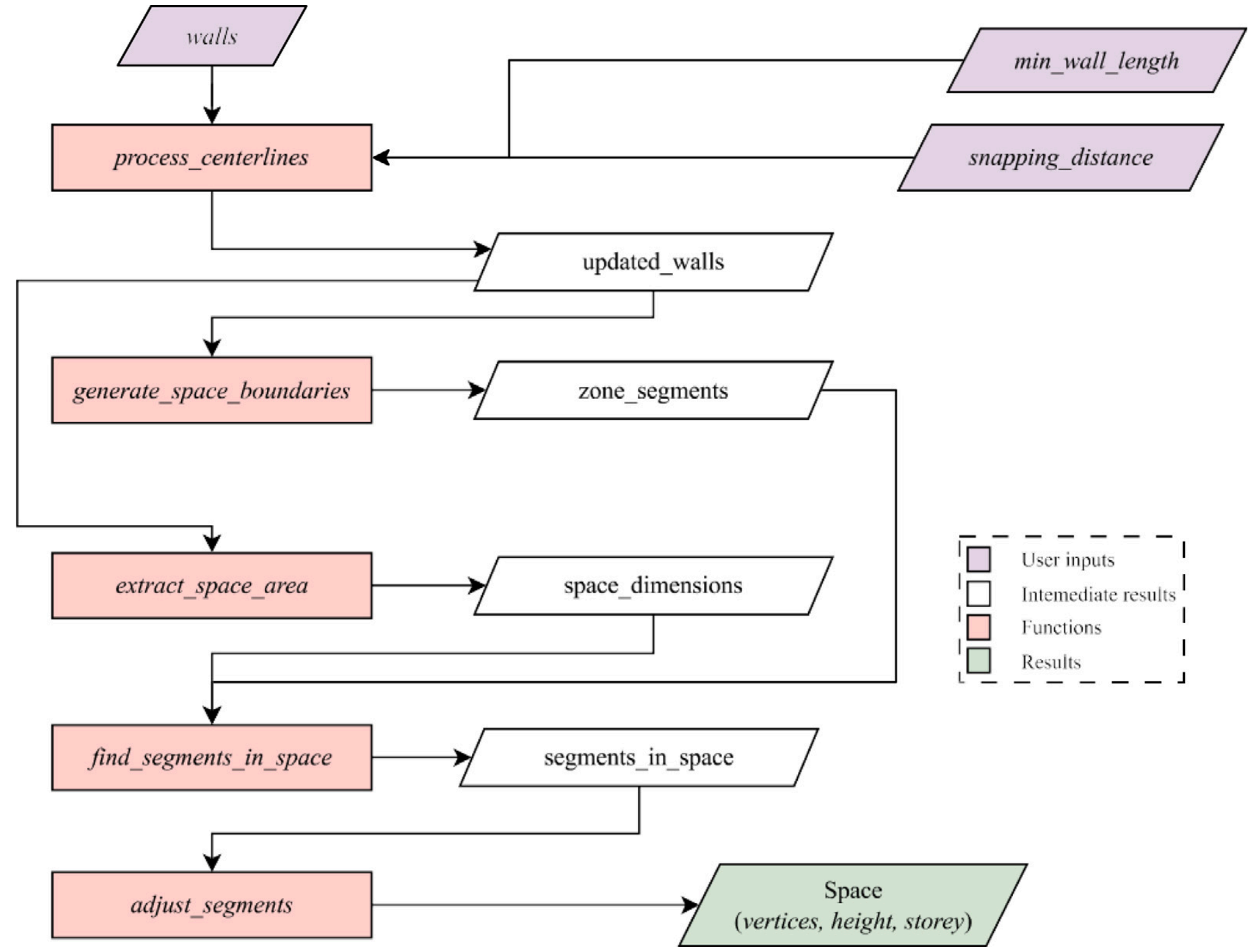

**Fig. 13.** Process of identifying zones from a floor plan.

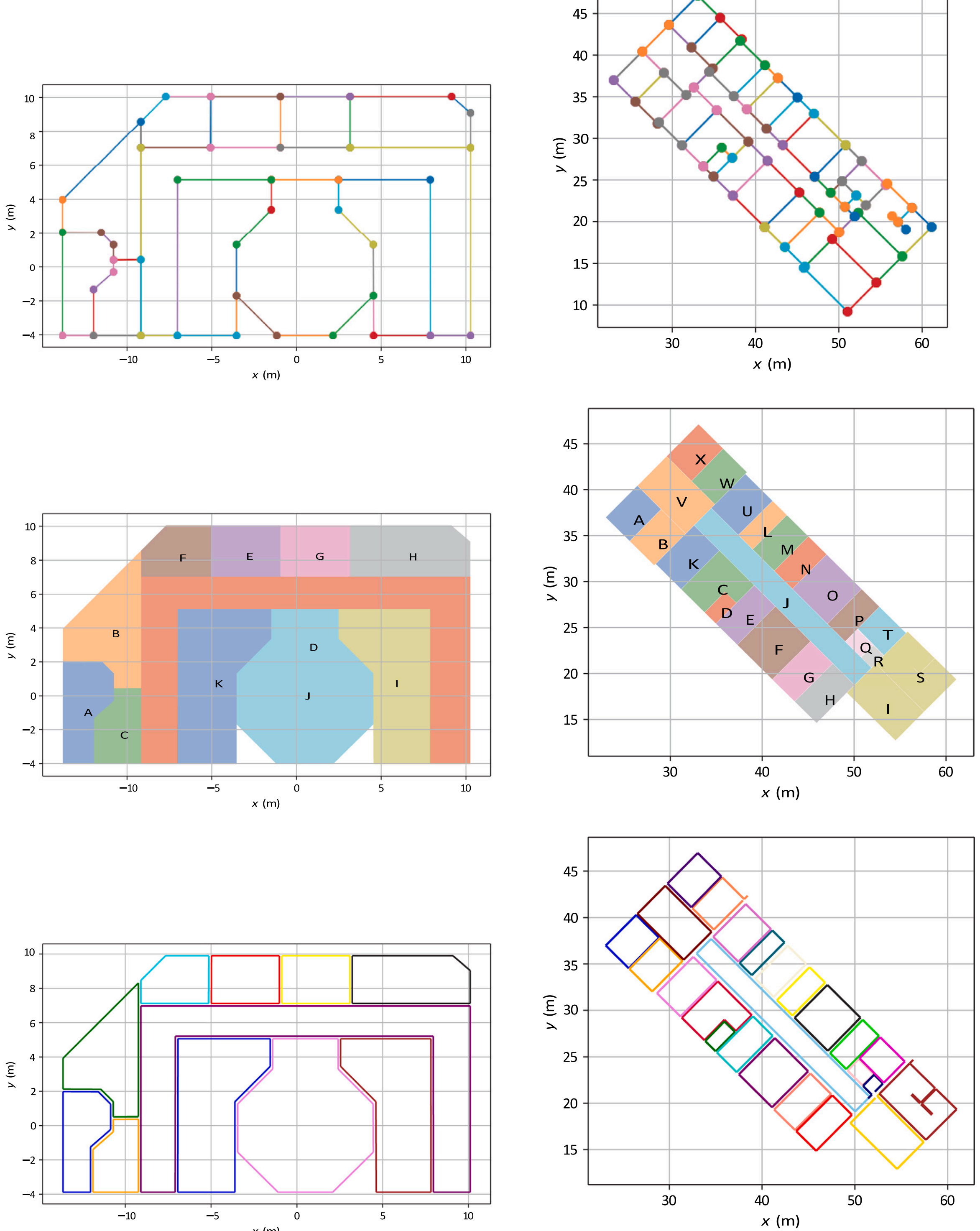

**Fig. 14.** Wall axes snapped to nearest neighbors and split at intersections (top), generated polygons based on wall axes (middle), and final zone boundaries defined by corresponding wall surfaces (bottom).

### 2.11. Inputs and calibration parameters

For accurate model configuration, Table 5 provides a comprehensive list of input parameters, encompassing details about the point cloud, the scanned building, and additional project information that cannot be derived directly from the laser scan. These parameters ensure that the software is precisely tailored to the specifics of each project. Table 6 details the calibration values essential for the effective classification and segmentation of building elements from the point cloud data, optimizing the accuracy of the generated model.

The calibration parameters in this study were selected based on empirical testing across multiple datasets to balance accuracy and computational efficiency. As Cloud2BIM evolves, a graphical user interface (GUI) will enable real-time adjustment of all parameters with a live preview of segmentation results, ensuring flexibility for diverse datasets.

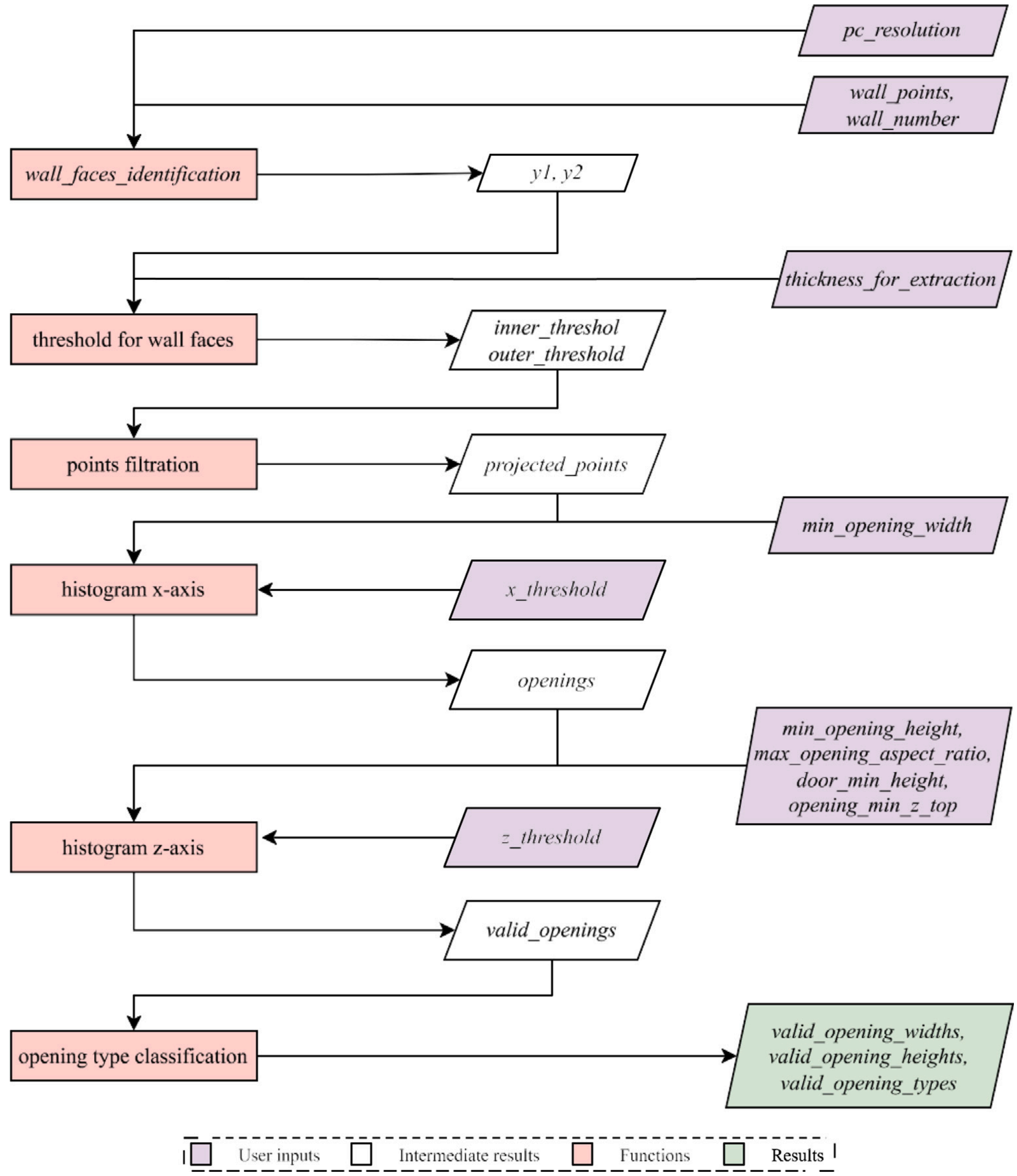

**Fig. 15.** Process of identifying openings from the distribution of points assigned to a wall.

For slab detection, *dilation_meters* and *erosion_meters* control morphological operations, compensating for occlusions such as furniture covering floors. These values must be fine-tuned based on dataset conditions. The *smoothing_factor*, regulating Douglas–Peucker simplification, remains minimal compared to overall element dimensions and typical point cloud measurement uncertainties. The *z_step* parameter, chosen heuristically, prevents peak loss in histograms while maintaining distinct slab surface separation, and *max_n_points_array* controls point density thresholds in slab detection.

For wall segmentation, *threshold* ensures all nonzero values contribute to mask generation, while *square(5)* determines kernel size in morphological processing. The *angle_tolerance* parameter, heuristically set, governs collinear segment merging. In opening detection, *max10* eliminates extreme peak values in histograms, with 10 chosen heuristically to filter out statistical outliers, though higher values could be considered.

## 3. Results and discussion

Cloud2BIM was evaluated through a series of carefully designed tests to highlight its performance and effectiveness in converting point clouds into BIM. These tests focused on two critical aspects: a comparative analysis to place our solution within the broader context of existing research and commercial tools, and a precision assessment using diverse datasets representative of typical use cases in the field.

### 3.1. Comparative analysis and benchmarking of algorithms

To accurately position Cloud2BIM within current developments, its capabilities were compared against existing methods. A detailed benchmarking of non-commercial solutions was conducted, highlighting their strengths and limitations across multiple key criteria and clarifying the advancements provided by Cloud2BIM. Additionally, commercial software solutions were evaluated to offer a comprehensive overview of the current state of point cloud conversion technologies.

#### 3.1.1. Non-commercial tools

To contextualize our software's capabilities within the field, Table 7 provides a comparative analysis of recent developments by other researchers tackling similar challenges in the point-cloud-to-BIM conversion. This table highlights essential criteria for classifying, segmenting, and reconstructing complete building point clouds, including the automation of multi-floor segmentation, the creation of volumetric elements like slabs, walls, and zones, and the handling of complex architectural features such as oblique wall axes and content elements like furniture. The criteria also cover the capabilities of opening detection, stairs modeling, and providing output in IFC file format, alongside the availability of the complete open-source software code. Comparative accuracies are not presented due to the lack of a uniform benchmark dataset and undisclosed accuracy metrics in many referenced studies.

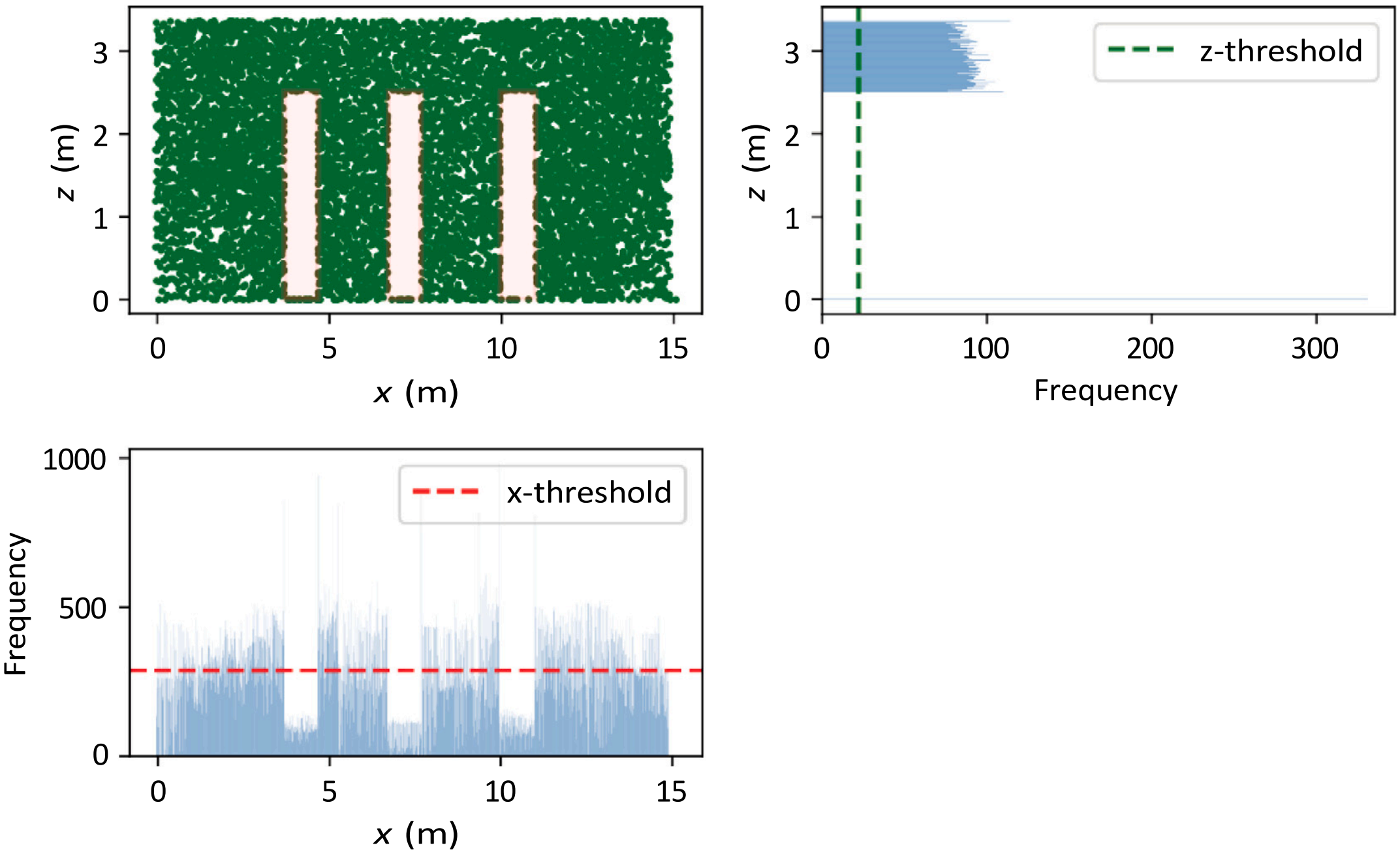

**Fig. 16.** Histogram-based wall opening detection showing the wall side view (top left), point density histograms along the $x$-axis (bottom left) and $z$-axis at the opening (top right), with dashed lines indicating detection thresholds.

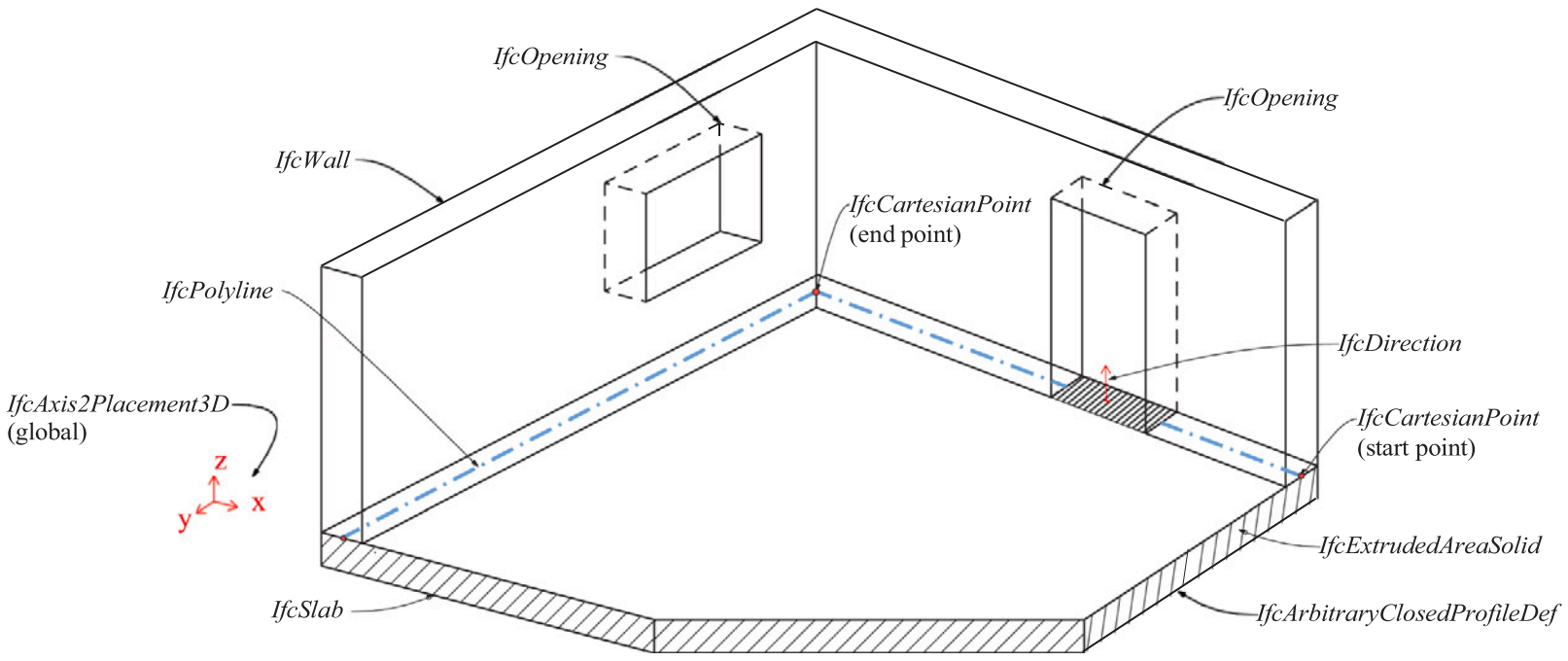

**Fig. 17.** Most important IFC classes used for geometry representation.

In reviewing the contributions of others, our discussion emphasizes the distinction in methodologies, scope, and outcomes. We highlight how different authors have approached the segmentation of point clouds and the geometric reconstruction of architectural elements, pointing out the strengths and limitations of their approaches compared to ours. For instance, Bassier and Vergauwen [19] focuses solely on walls, creating geometry in Revit through a Rhino plug-in and then exporting to IFC, including zones. Our method supports direct IFC export from Python, reducing dependence on commercial software. In addition, their approach, like some others, lacks the ability to detect and model doors and windows, often resulting in a lower level of detail.

The work of Mahmoud et al. [11] incorporates advanced deep learning techniques for semantic segmentation, followed by RANSAC for line extraction to identify architectural features. While it provides detailed segmentation of walls, floors, and ceilings, the geometry export is limited to an Excel format, which is then used by a Dynamo add-in for Revit, constraining the output to specific commercial software. Similarly, the work by Mehranfar et al. [30] relies on the Revit API for 3D geometry and employs the PointTransformer network with DBScan clustering, also utilized in an improved version by Czerniawski et al. [54]. However, this approach is sensitive to auxiliary parameter settings and does not enhance functionality beyond our method. Furthermore, it only supports single-floor point clouds, limiting its utility for multi-storey buildings.

Martens and Blankenbach [10] presented VOX2BIM+, a robust approach by analyzing voxels to improve noise resistance and accuracy in the volumetric reconstruction of spaces and structural elements. Unlike our approach, which utilizes direct point cloud processing, their method incorporates several advanced techniques, including region growing [55] and morphological skeletons for wall detection. However, their code remains proprietary, and their method does not support window or door geometry extraction.

Similarly, Wang et al. [9] and Ochmann et al. [18] have made significant contributions to the field, focusing on energy simulation models to detailed wall segmentation. These studies typically rely on simplified models or specific algorithms like RANSAC, which may only capture some architectural details such as doors and windows, a capability our software aims to address comprehensively. While sufficient for their goals, the detected zones that are clustered according to their purpose and not published to IFC cannot be further edited and exported to other software.

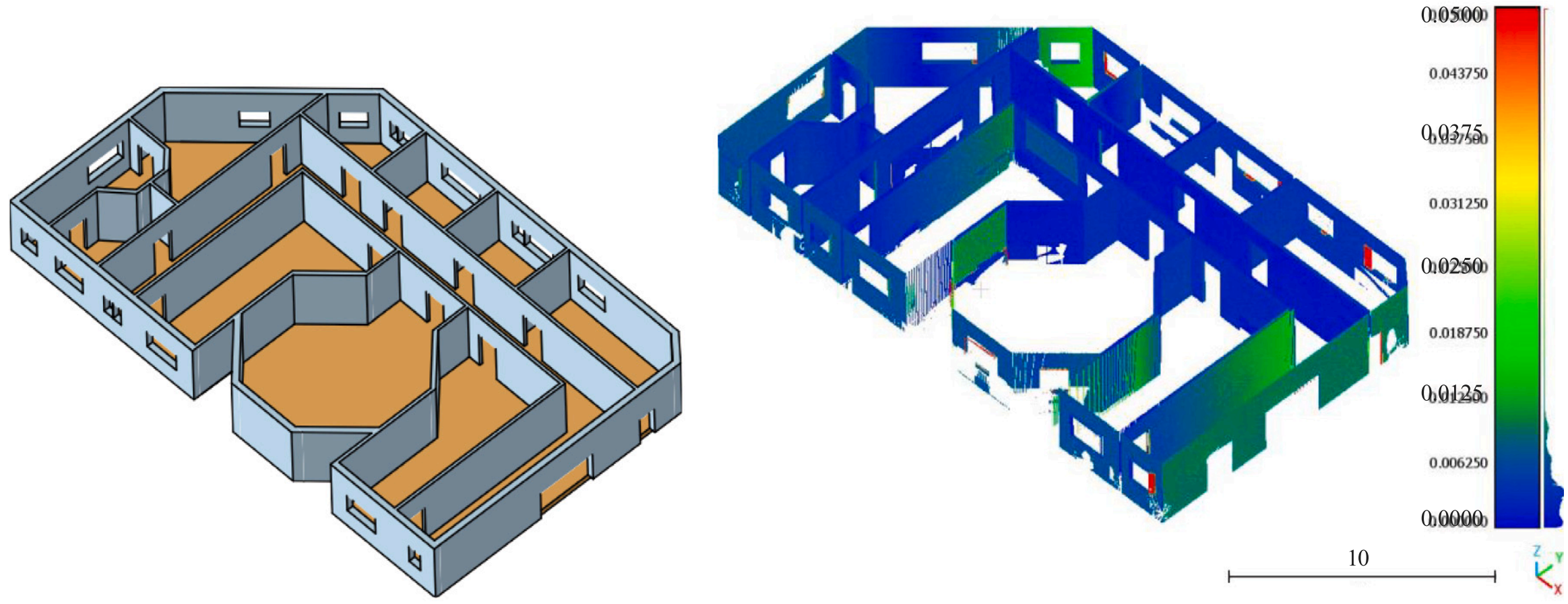

**Fig. 18.** IFC model of the ETH Zurich synth3 dataset, automatically generated in Cloud2BIM (left) and a heat map of absolute distances (m) between the point cloud and the resulting IFC model (right).

**Table 5**
Input parameters for the Cloud2BIM software, detailing settings related to the point cloud, building elements, and project details.

| Parameter | Unit | Related object/entity |
|---|---|---|
| *pc_resolution* | m | Point cloud |
| *bfs_thickness* | m | Slab |
| *tfs_thickness* | m | Slab |
| *min_wall_length* | m | Wall |
| *min_wall_thickness* | m | Wall |
| *max_wall_thickness* | m | Wall |
| *exterior_walls_thickness* | m | Exterior wall |
| *snapping_distance* | m | Wall axis, zone |
| *material_for_objects* | – | Material |
| *ifc_site_latitude* | degree | Site |
| *ifc_site_longitude* | degree | Site |
| *ifc_site_elevation* | m | Site |
| *ifc_project_name* | – | Project |
| *ifc_project_long_name* | – | Project |
| *ifc_project_version* | – | Project |
| *ifc_author_name* | – | Author |
| *ifc_author_surname* | – | Author |
| *ifc_author_organization* | – | Author |
| *ifc_building_name* | – | Building |
| *ifc_building_type* | – | Building |
| *ifc_building_phase* | – | Building |

Often very advanced approaches, employed assumptions limiting the reconstructed structures to orthogonal layouts or single story structures, but their findings were useful for Cloud2BIM development [52, 53].

### 3.1.2. Commercial tools

To provide additional context on commercial solutions in the point-cloud-to-BIM domain, Table 8 summarizes key functionalities of various tools. The table highlights the presence of semantic segmentation capabilities, the ability to detect walls and openings, multi-story support, IFC output, and integration with Autodesk Revit.

Several commercial solutions focus on semantic segmentation of point clouds, identifying elements such as walls, floors, and openings (e.g., Pointfuse, Aurivus, Faramoon, and Geo-Plus VisionBIM). However, their approach to BIM conversion varies significantly. Pointfuse/Autodesk Recap provides semantic segmentation and basic geometric modeling but lacks robust multi-story automation. Aurivus specializes in semantic classification of point clouds without generating BIM models directly, instead offering a Revit plugin for further modeling. Faramoon generates an initial IFC model at LOD100, which is then manually refined. Geo-Plus VisionBIM automates part of the modeling process, including AI-assisted window detection, but still requires substantial user input to define structural parameters.

IFC export is available in Pointfuse, Faramoon, usBIM.Scan2IFC, and Geo-Plus VisionBIM. However, many solutions do not generate fully structured BIM elements and instead rely on user intervention within proprietary software. usBIM.Scan2IFC offers multi-story support, but the user must manually define height levels, limiting automation.

Faramoon operates as a black-box service, where users submit point clouds, and the company returns a modeled BIM representation. This contrasts with tools like Undet and Leica CloudWorx, which function as Revit plugins that assist users in manually aligning and placing elements rather than offering full automation. These tools employ a fit-based approach, where the user clicks on areas where elements should be placed, and the software adjusts the positioning accordingly.

ClearEdge3D EdgeWise and BricsCAD BIM v24.0 incorporate additional automation features. However, ClearEdge3D EdgeWise struggles with inaccurate wall connections and lacks opening detection, making its outputs less consistent. Multi-story support remains limited across most solutions, often requiring floor-by-floor processing. While BricsCAD BIM v24.0 allows users to classify point clouds and generate zones, it does not detect openings or support fully automated multi-story modeling. In contrast, Geo-Plus VisionBIM does support multi-story workflows, though with significant manual input required to refine the detected elements.

## 3.2. Precision analysis

This section evaluates the precision of the segmentation, classification, and reconstruction algorithms implemented in Cloud2BIM using the datasets introduced in Section 2.1. Although established metrics exist for quantifying the precision of 3D model reconstructions [56], these were not employed in our evaluation due to their dependency on manually created ground-truth models, which were not available. Instead, we present visual assessments complemented by heat maps illustrating absolute distances between the original point clouds and the resulting IFC model surfaces. These visualizations effectively demonstrate the achieved precision of the proposed methods across all tested datasets.

### 3.2.1. ETH Zurich synth3

The ETH Zurich synth3 dataset was selected to showcase the performance of our volumetric method in wall segmentation within a 2.5-dimensional Atlanta-world system. This dataset is frequently used to benchmark the success of interior wall segmentation [18,19,38]. Although it lacks an exterior scan, this limitation mirrors real-life scenarios where facades are unscanned, thus providing a relevant challenge. Our algorithm successfully classified and generated the geometry for all interior walls, with exterior walls modeled using a predefined thickness. Fig. 18 illustrates the resulting IFC model, with the upper ceiling slab hidden to enhance clarity. This model accurately reflects the volumes, positions of openings, and room dimensions. Fig. 18 also shows the absolute distance from the points to the wall surfaces, typically within millimeters; a similar accuracy with the deviations up to 2.5 mm in case of walls was achieved also by [38]. In contrast, the VOX2BIM+ algorithm [10] exhibited significantly more significant deviations.

**Table 6**
Calibration parameters for the segmentation and classification algorithms, including default values and descriptions for each setting to optimize model accuracy; the default values were set based on testing the algorithms based on different datasets, but in the near future the software will be equipped with sliders and graphical interface to allow convenient setting of these parameters.

| Parameter | Unit | Default value | Description |
|---|---|---|---|
| *dilution_factor* | – | 10 | Controls the skipping of rows in the XYZ point cloud file. |
| *grid_coefficient* | px/mm | 5 | Sets the grid size for computational operations in binary image processing. |
| *z_step* | m | 0.05 | Defines the bin size for the horizontal histogram in slab surface detection. |
| *max_n_points_array* | – | 0.5 | Multiplier for maximum point density to identify $z$-coordinates for slabs. |
| *dilation_meters* | m | 1 | Specifies the dilation distance for morphological operations on slab masks. |
| *erosion_meters* | m | 1 | Specifies the erosion distance for morphological operations on slab masks. |
| *smoothing_factor* | m | 0.0005 | Determines the smoothing level of slab polygon contours using the Douglas-Peucker algorithm. |
| *safety_margin* | m | 0.1 | Margin (overlap) used when splitting the point cloud into storeys. |
| *z_section_boundaries* | – | [0.9, 1.0] | Sets the height percentage thresholds for segmenting point clouds for wall detection. |
| *threshold* | – | 0.01 | Sets the value threshold for creating a positive mask from 2D histograms. |
| *square(5)* | px | 5 | Kernel size and shape for adjusting the wall detection mask. |
| *epsilon* | m/px | 0.02 | Controls the approximation of line segments. |
| *angle_tolerance* | degree | 3 | Tolerance level for merging collinear segments during wall reconstruction. |
| *max10* | – | 10 | Sets the threshold for determining if a histogram bin represents an opening. |

**Table 7**
Comparative analysis of the functionality of solutions reported in recent articles focused on the point cloud to model conversions. Both recent and older studies are included to provide a comprehensive overview, reflecting the ongoing and historical contributions to this specialized field.

| | Multi-story[a] | Slabs[b] | Non-Manhattan walls[c] | Wall surfaces[d] | Volumetric walls[e] | Volumetric spaces[f] | Openings[g] | IFC output[h] | Open-source[i] |
|---|---|---|---|---|---|---|---|---|---|
| Wang et al. [9] | ✘ | ✘ | ✓ | ✓ | ✘ | ✘ | ✓ | ✘ | ✘ |
| Macher et al. [27] | ✓ | ✓ | ✓ | ✓ | ✓ | ✘ | ✘ | ✓ | ✘ |
| Jung et al. [20] | ✘ | ✘ | ✓ | ✓ | ✓ | ✘ | ✓ | ✘ | ✘ |
| Shi et al. [38] | ✘ | ✘ | ✓ | ✓ | ✓ | ✘ | ✓ | ✘ | ✘ |
| Ochmann et al. [18] | ✓ | ✓ | ✓ | ✓ | ✓ | ✓ | ✘ | ✘ | ✘ |
| Bassier and Vergauwen [19] | ✓ | ✘ | ✓ | ✓ | ✓ | ✓ | ✘ | ✓ | ✓ |
| Romero-Jarén and Arranz [8] | ✓ | ✘ | ✓ | ✓ | ✘ | ✘ | ✘ | ✘ | ✘ |
| Gen and Gentes [52] | ✘ | ✘ | ✘ | ✓ | ✓ | ✘ | ✘ | ✘ | ✘ |
| Martens and Blankenbach [10] | ✓ | ✓ | ✓ | ✓ | ✓ | ✘ | ✘ | ✓ | ✘ |
| Mahmoud et al. [11] | ✘ | ✓ | ✓ | ✓ | ✘ | ✘ | ✓ | ✘ | ✘ |
| Mehranfar et al. [53] | ✓ | ✘ | ✘ | ✓ | ✓ | ✓ | ✘ | ✘ | ✘ |
| Our solution (Cloud2BIM) | ✓ | ✓ | ✓ | ✓ | ✓ | ✓ | ✓ | ✓ | ✓ |

[a] Automatic processing of multi-story buildings, where floors are detected and structured hierarchically (e.g., as IfcBuildingStorey) without requiring manual intervention.
[b] Detection of slab position, thickness, and horizontal dimensions.
[c] Support for walls that are not strictly aligned to an orthogonal (Manhattan) grid, allowing for arbitrary angles.
[d] Identification and semantic segmentation of wall surfaces from point clouds.
[e] Detection of walls as solid objects, including their topology (connections between walls) and representation as parametric elements with thickness, start and end points, and height.
[f] Creation of volumetric room representations, which are derived separately from wall geometry and are not necessarily included in solutions that detect volumetric walls.
[g] Geometric representation and placement of openings within wall elements, including width, height, position, and type (door, window).
[h] Automatic conversion of extracted building geometry into the IFC format without reliance on proprietary software.
[i] The tool's code is publicly available, allowing free use and modification.

The dataset also facilitated testing our histogram-based technique complemented by heuristic rules for detecting openings. All internal doors were accurately identified (13/13), although the detection was occasionally compromised by furniture or large gaps in point density on the walls. Such errors require manual corrections during model editing in any BIM software (see Section 3.4, Fig. 25).

### 3.2.2. Kladno station

The Kladno station dataset was used to validate our wall segmentation algorithm on an actual building rather than synthetic data, which was challenging due to the presence of furniture, wall curvatures, and atypical structural elements, such as wall openings with bars. This wall did not conform to standard thickness parameters, highlighting the algorithm's limitations in processing non-standard elements. The resulting IFC model is depicted in Fig. 19. This figure also demonstrates the absolute distance of points from the surfaces, with deviations up to 25 mm. The heat map highlights points over 50 mm away, often associated with staircases, windows, or furnishing elements. These elements were not identified and reproduced, yielding high inaccuracies in the heat map. When detecting zones, one room was incorrectly merged with the other due to failed snapping of imperfectly built walls in a corner of the building, resulting in 24/25 (96%) accuracy (recall Fig. 14).

### 3.2.3. Prague hotel

The Prague hotel dataset, a comprehensive scan of a former hotel, showcased our capability to segment extensive building datasets and underscored the accuracy of our ceiling slab segmentation. The model, displayed with transparent walls for clarity, highlights ceiling slabs in yellow (Fig. 20). The dataset posed challenges in slab identification due to obstacles and variable point densities, particularly at ground level, where furniture heavily influenced results. Slabs, especially those with only one surface visible, required precise calibration of morphological operations to ensure accurate segmentation. The absolute distances from the point cloud to the slab surfaces, shown in Fig. 20, vary up to 25 mm. A few places exhibit the error exceeding 50 mm due to missing point cloud data for the area.

### 3.2.4. ISPRS benchmark datasets

Despite achieving satisfactory reconstruction results, we did not use the ISPRS benchmark datasets [57] to evaluate the accuracy of Cloud2BIM. These datasets, designed to support research in automated 3D indoor modeling, include predefined evaluation metrics based on completeness, correctness, and accuracy. However, these metrics require a manually created ground-truth model, which introduces potential inaccuracies. Furthermore, we identified several obstacles in these datasets, including occlusions, clutter, and non-standard architectural features, making them less suitable than those selected in Section 2.1 for assessing Cloud2BIM's performance. Since we already had dedicated benchmark datasets, no additional ones were included to maintain the study's focus. However, Cloud2BIM is freely available [32], and readers are encouraged to conduct their own evaluations using the ISPRS datasets.

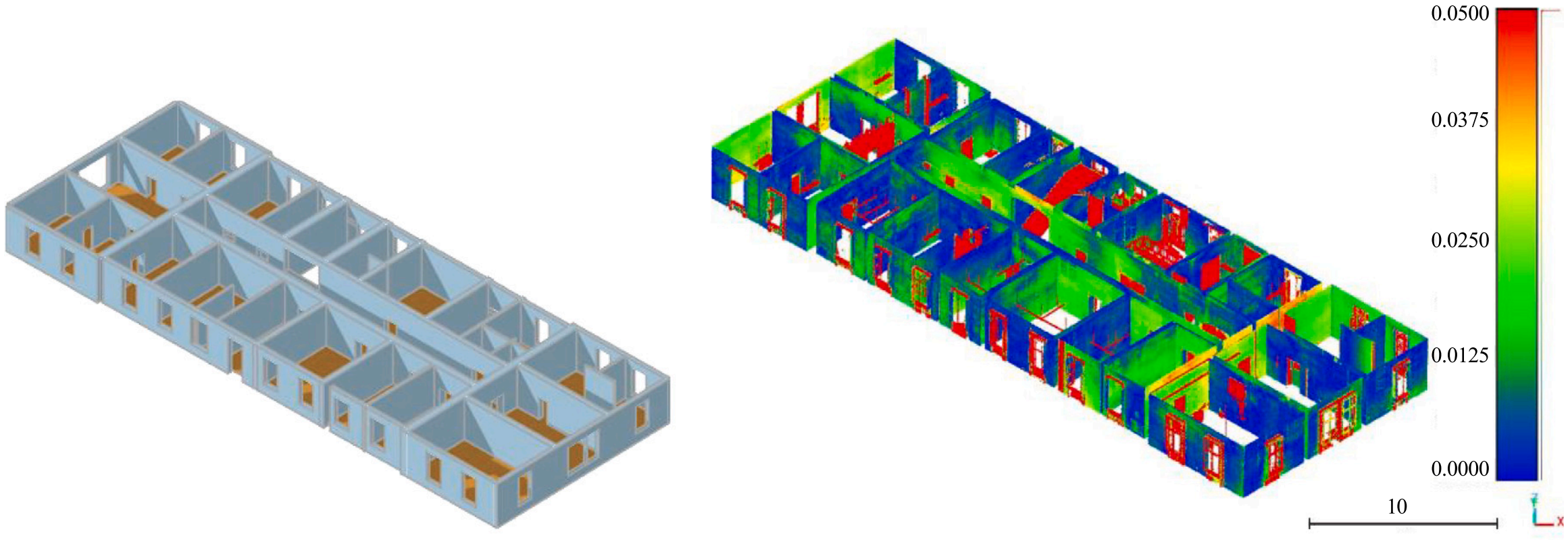

**Fig. 19.** IFC model of the Kladno station dataset, automatically generated in Cloud2BIM (left) and a heat map of absolute distances (m) between the point cloud and the resulting IFC model (right).

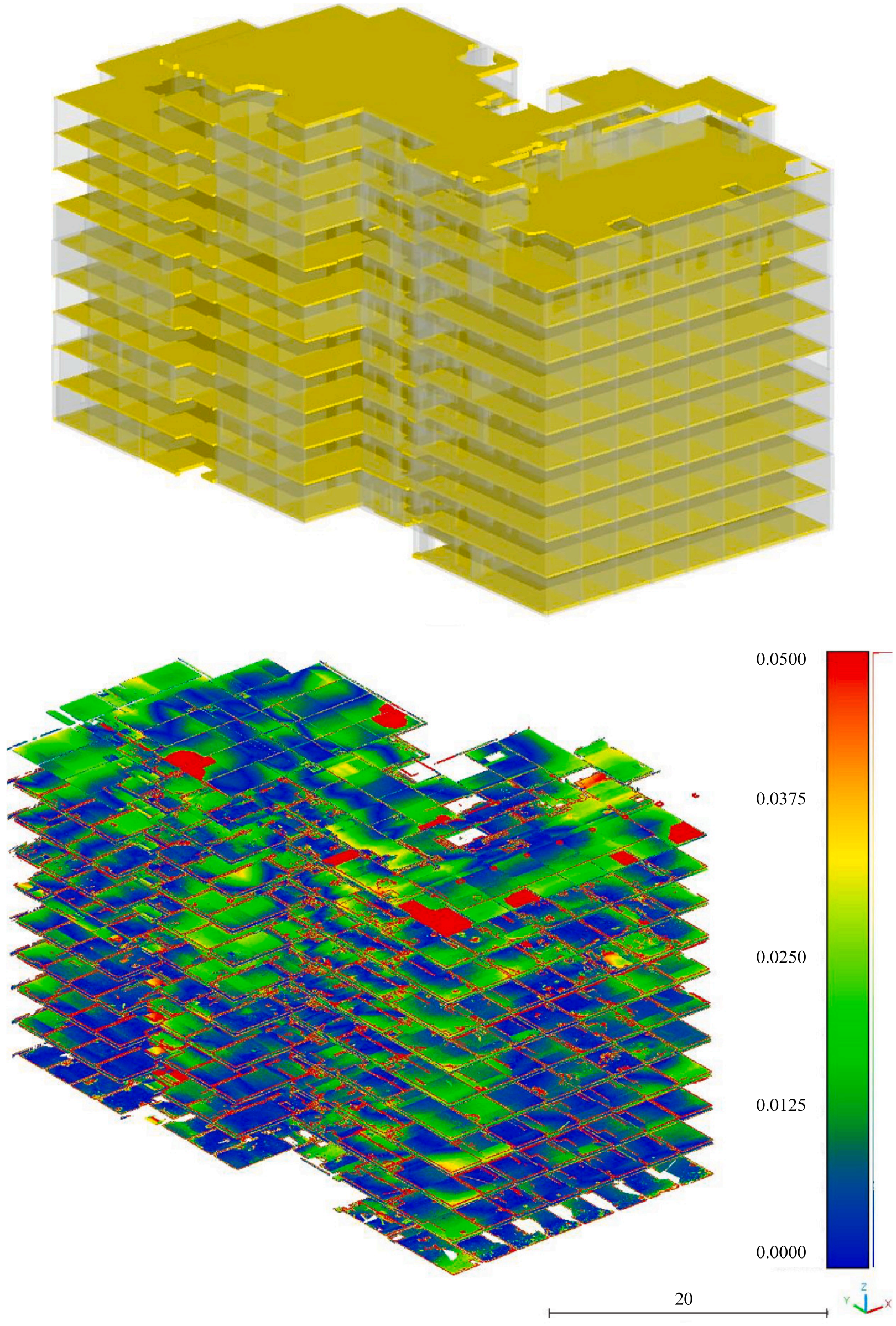

**Fig. 20.** IFC model of the Prague hotel dataset with slabs highlighted in yellow, automatically generated in Cloud2BIM (top) and a heat map of absolute distances (m) between the point cloud and the resulting IFC model (bottom).

**Table 8**
Overview of commercial tools for point cloud to BIM conversion, summarizing their key functionalities.

| Tool | Semantic segmentation | Walls and openings detection | Multi-story support | IFC output |
|---|---|---|---|---|
| Pointfuse/Autodesk Recap | ✓ | ✓ | ✓[a] | ✓ |
| Aurivus | ✓ | ✗ | ✓ | ✗ |
| Faramoon | ✓ | ✓ | ✓ | ✓ |
| usBIM.Scan2IFC | ✗ | ✓ | ✓[b] | ✓ |
| Undet | ✗ | ✓[c] | ✗ | ✗ |
| ClearEdge3D EdgeWise | ✓ | ✓[d] | ✗ | ✗ |
| Geo-Plus VisionBIM | ✓ | ✓ | ✓ | ✓ |
| BricsCAD BIM v24.0 | ✓ | ✓[e] | ✗ | ✗ |
| Leica CloudWorx | ✗ | ✓[f] | ✗ | ✗ |

[a] Exports to IFC at LOD100, requiring manual refinement.
[b] Multi-story support is available, but users must manually define height levels.
[c] Wall and opening detection requires manual selection (Fit function).
[d] Walls are detected, but openings are not. Wall detection accuracy is limited.
[e] Wall detection is available, but openings are not detected.
[f] Wall and opening placement is assisted but requires manual selection.

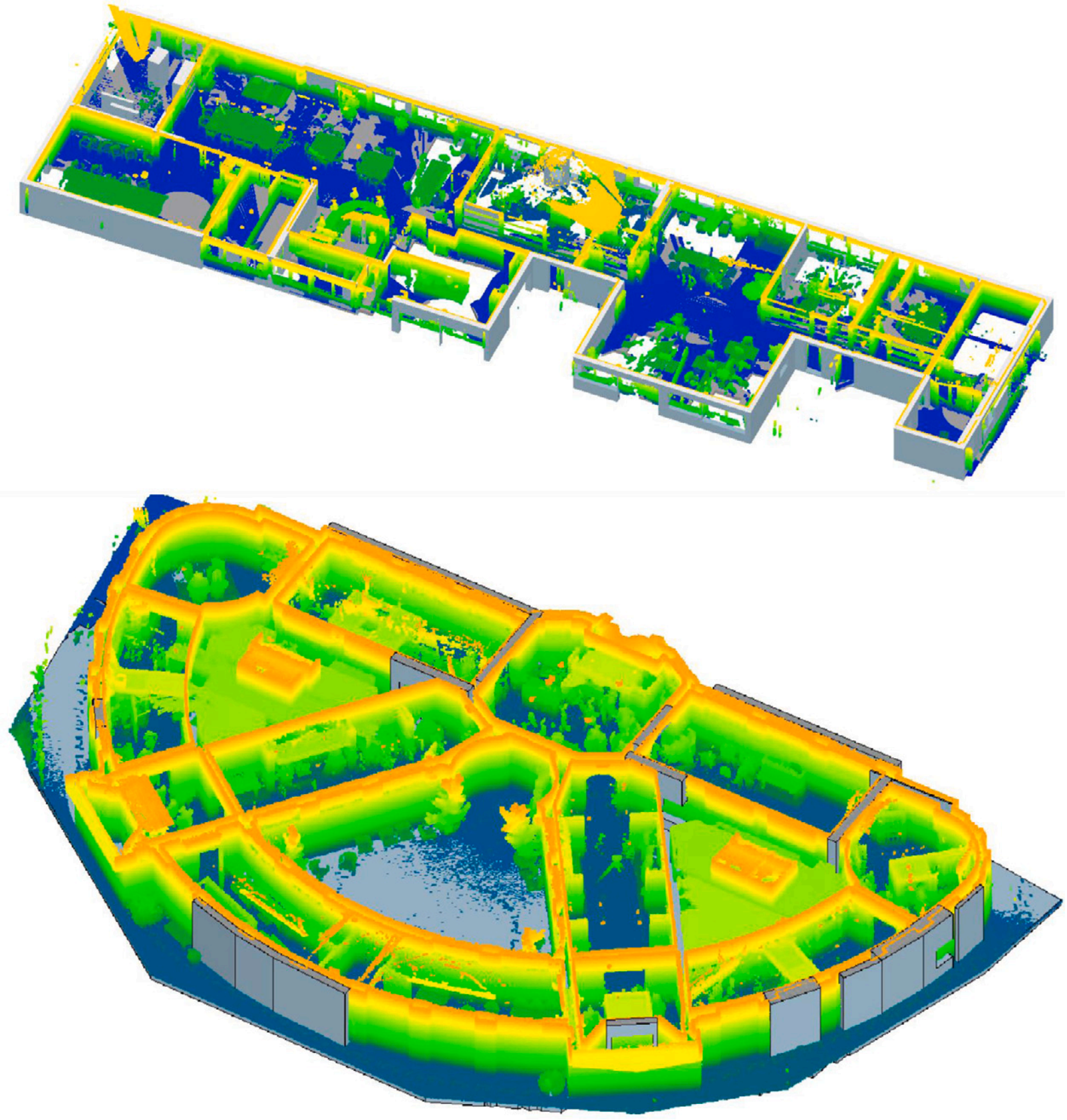

**Fig. 21.** Reconstruction results for the ISPRS Fire Brigade (top) and Grainger Museum (bottom) benchmark datasets; the point clouds are color-coded by height (blue to red), while the reconstructed IFC elements, including slabs, walls, and detected openings, are displayed in gray.

To illustrate the characteristics of the ISPRS benchmark datasets, we provide an overview of individual point clouds along with figures demonstrating their reconstruction using Cloud2BIM:

- Fire Brigade (Delft, The Netherlands): This dataset was acquired using a terrestrial laser scanner and is characterized by a high level of clutter, as noted by the dataset authors. Structural elements are partially obstructed by furniture and other objects, and occlusions in static laser scanning result in data gaps. Despite these challenges, Cloud2BIM successfully reconstructed external walls and detected most openings (Fig. 21, top image). The indoor scene contains 9 rooms, 10 doors, and 53 windows, with additional complexity introduced by curtain walls containing windows.
- Grainger Museum (Melbourne, Australia): This dataset represents a *Non-Manhattan World* building with 18 rooms, 13 doors, and 41 windows. Captured using the Zeb Revo RT system from both indoors and outdoors, it features significant clutter due to the presence of various artifacts. Curved walls posed a major challenge, as Cloud2BIM is designed for rectilinear structures. When detected, these curved walls were approximated using straight

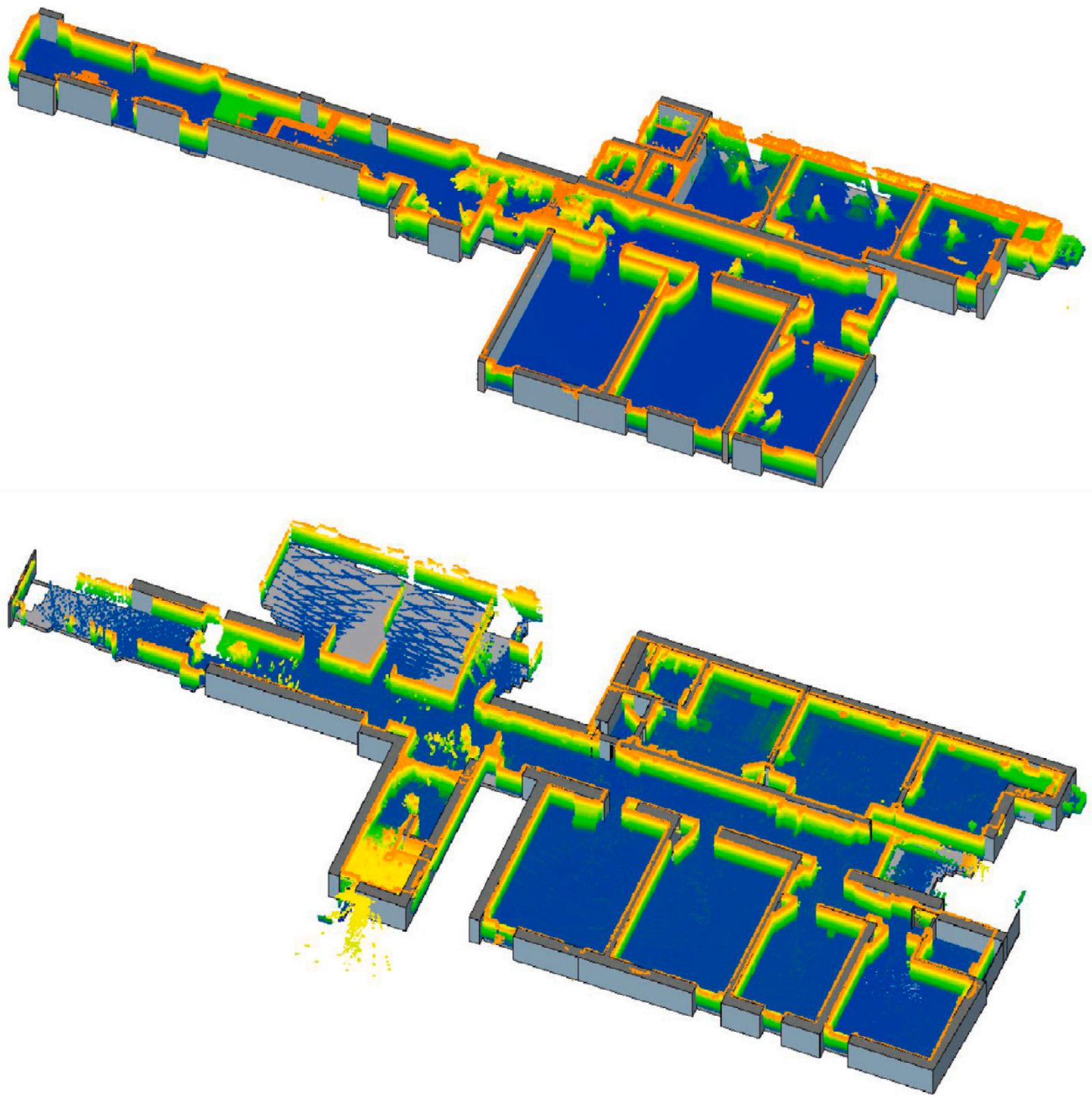

**Fig. 22.** Reconstruction results for the ISPRS TUB1 (top) and TUB2 (bottom) benchmark datasets; the point clouds are color-coded by height (blue to red), while the reconstructed IFC elements, including slabs, walls, and detected openings, are displayed in gray.

segments, highlighting a limitation of the current workflow (Fig. 21, bottom image).

- TUB1 (Technical University of Braunschweig, Germany): Captured with the Viametris iMS3D system, this dataset includes both a point cloud and the sensor trajectory. It features 10 rooms enclosed by walls of varying thicknesses, 23 doors (open and closed), and 7 windows. As the building is unfurnished, the clutter level is low, primarily consisting of points corresponding to people present during the survey. Cloud2BIM handled this dataset well, although variable floor levels posed challenges, requiring careful tuning of input parameters (Fig. 22, top image).
- TUB2 (Technical University of Braunschweig, Germany): Acquired in the same building as TUB1 but using a Zeb-Revo scanner, this dataset covers two floors connected by a staircase. The first level contains 14 rooms, 8 windows, and 23 doors, while the second level consists of 10 rooms, 13 windows, and 28 doors. Wall thicknesses and ceiling heights vary. Like TUB1, the dataset has low clutter, making it a relatively straightforward test case for Cloud2BIM (Fig. 22, bottom image).

Following this overview, we conducted an additional evaluation using the TUB2 dataset to provide a more detailed analysis as provided by Khoshelham et al. [58]. However, when attempting to quantify the accuracy of Cloud2BIM's output by comparing it with the original IFC model provided in the dataset, we observed significant discrepancies. Specifically, the alignment between the original IFC model and the reconstructed model was poor, with deviations between corresponding wall surfaces reaching up to 10 cm (Fig. 23). This discrepancy resulted in a notably low intersection-over-union (IoU) value when comparing the two models (Fig. 24).

Upon detailed inspection, we found that the original IFC model itself does not accurately correspond to the provided point cloud, introducing errors that compromise the reliability of any direct comparison. These findings support the argument that evaluation metrics relying on manually created ground-truth models introduce uncertainty and may not provide a fully reliable basis for assessing reconstruction precision.

Nonetheless, we acknowledge that the ISPRS datasets provide valuable insights into the robustness of reconstruction algorithms under challenging conditions, such as occlusions, clutter, and non-standard architectural features. Therefore, these datasets, while not suitable for precise quantitative evaluation of reconstruction accuracy, remain highly valuable for testing algorithm robustness and capabilities in complex environments.

### 3.3. Speed performance analysis

This subsection presents the computational performance results for the test datasets described in Section 2.1. The evaluations were conducted on a Dell Precision 3591 notebook with an Intel Core Ultra 165H CPU (1.40 GHz), 32 GB RAM, a PM981 NVMe Samsung SSD, and a Windows 11 operating system. The software, developed in Python 3.10, was tested with datasets chosen for their distinctive properties, as detailed in Section 2.1. Table 9 summarizes the computation times alongside parameters of the resulting models and their point clouds.

While Python's interpreted nature might not traditionally align with high-performance applications compared to compiled languages like C# or C++, it offered significant advantages in terms of development speed. We leveraged data vectorization capabilities provided by NumPy arrays to boost performance, which proved particularly effective for smaller datasets. For example, the entire reconstruction of the Kladno station, which is relatively modest in scale, was completed in just 30 seconds on standard hardware specifications. In contrast, a more extensive dataset like the Prague hotel, comprising over 40 million points, required 30 min to process. This increase in processing time is mainly due to the complexity involved in pairing and joining numerous wall segments. This task could benefit from a compiled language implementation once the software's development reaches maturity.

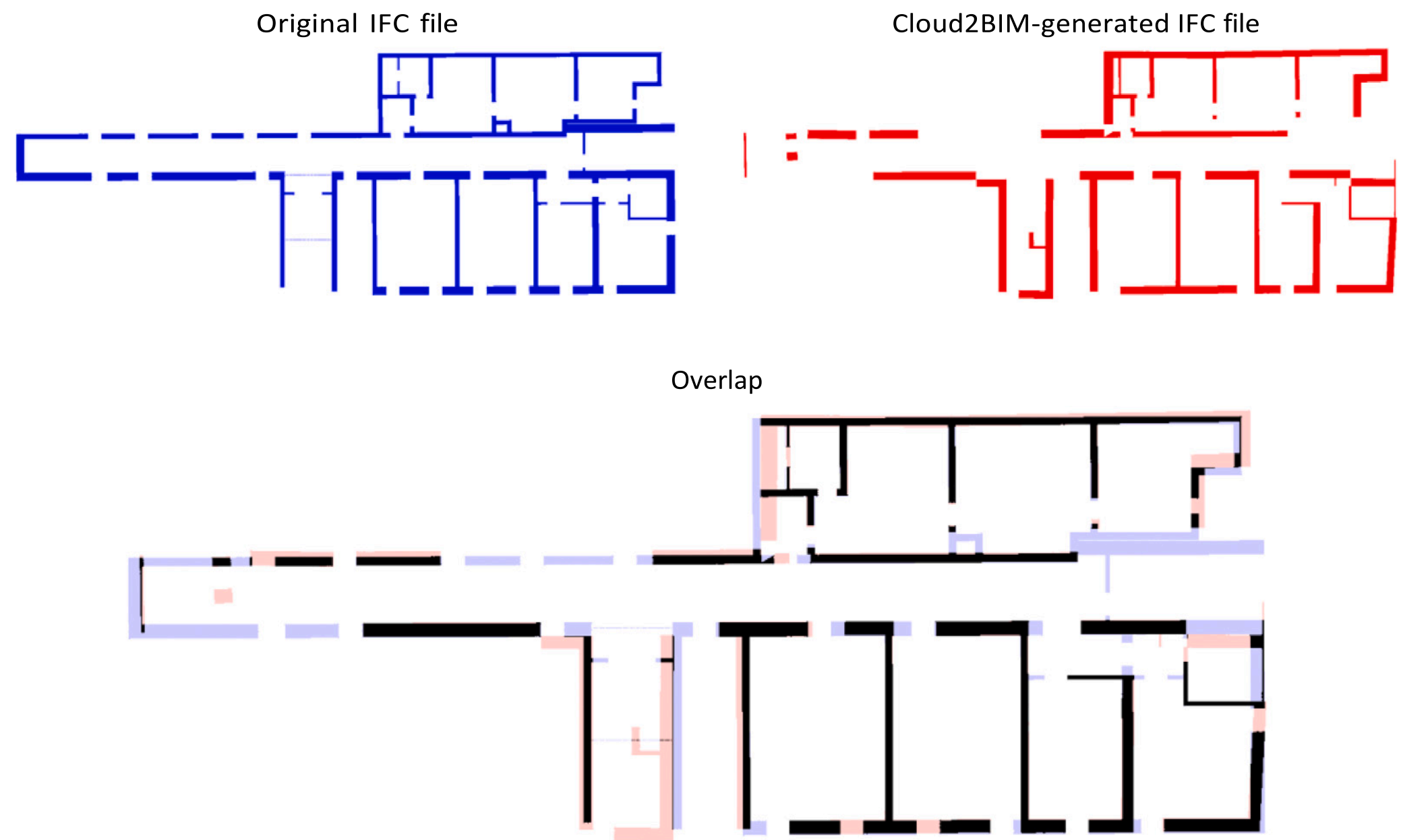

**Fig. 23.** IoU between the original IFC model of the ISPRS TUB2 dataset and the reconstructed IFC model using Cloud2BIM based on the provided point cloud; the IoU value reached 0.47.

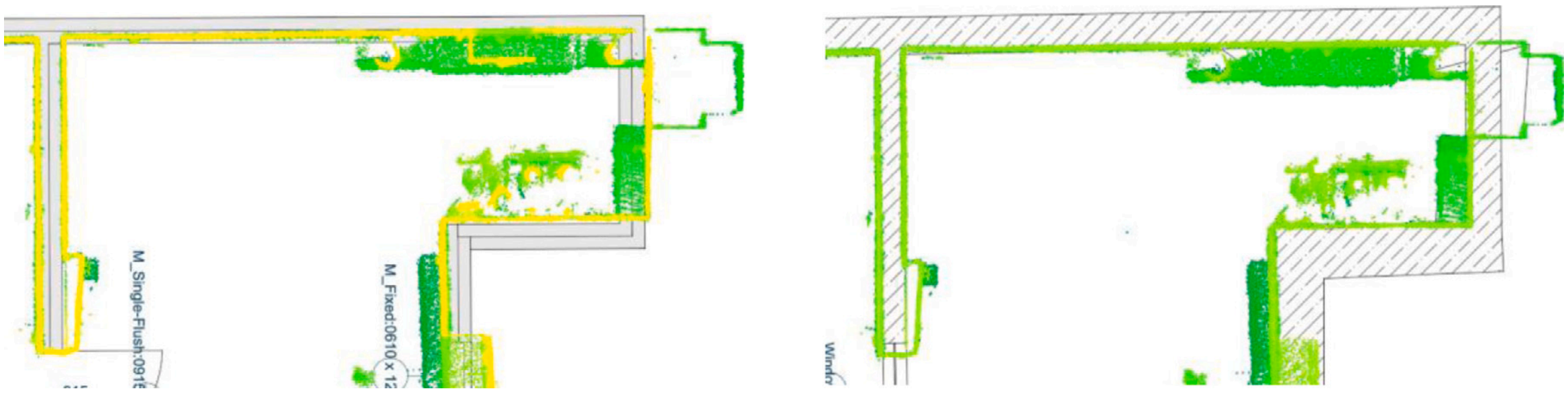

**Fig. 24.** Original IFC model for the ISPRS TUB2 dataset shows discrepancies from the point cloud, reaching up to 10 cm (left), whereas the reconstructed IFC model using Cloud2BIM matches the point cloud (right) and therefore differs from the original IFC file.

**Table 9**
Overview of computation results for the test datasets.

| Dataset | Points | Floor area | Number of elements | Computation time |
|---|---|---|---|---|
| ETH Zurich synth3 | 6,905,684 | 303 $m^2$ | 74 | 60 s |
| Kladno station | 2,551,721 | 571 $m^2$ | 109 | 30 s |
| Prague hotel | 40,152,849 | 9348 $m^2$ | 621 | 30 min |

Despite its current developmental state, Cloud2BIM outperforms similar software solutions, as evidenced by a comparative analysis with the work by Macher et al. [27], who did not incorporate opening identification and have not released their code as open-source. Our software demonstrated approximately 7× the processing efficiency: in a benchmark test, our system processed a 7 million point dataset at a rate of 6.9 million points per minute, compared to the 1.25 million points per minute achieved by the aforementioned study on a 10 million-point dataset. For context, their system required 8 minutes to process a house with approximately 10 million points covering 200 $m^2$ and 19 min for the main block of an office building encompassing 1000 $m^2$ with approximately 16 million points (Table 7). Shi et al. [38] also used the ETH Zurich synth3 dataset and its processing took them 359 s without providing the IFC output while we reached the total processing time, including IFC generation, of 60 s (Table 9).

### 3.4. Limitations

Despite the promising results achieved in this study, several limitations must be acknowledged. These limitations highlight areas for future research and potential enhancements to further improve the robustness and applicability of the proposed method.

First, the method does not currently support the detection and reconstruction of curved walls. The implemented algorithm is optimized for planar geometries, meaning that curved surfaces are either approximated by segmented planar sections or may not be processed correctly. This limitation is particularly relevant for buildings with complex architectural features, such as circular walls, domed structures, or organic geometries.

Second, the system is currently limited to detecting and reconstructing rectangular and square window and door openings. While these shapes encompass the majority of standard architectural designs, they do not account for arched, circular, or irregularly shaped openings.

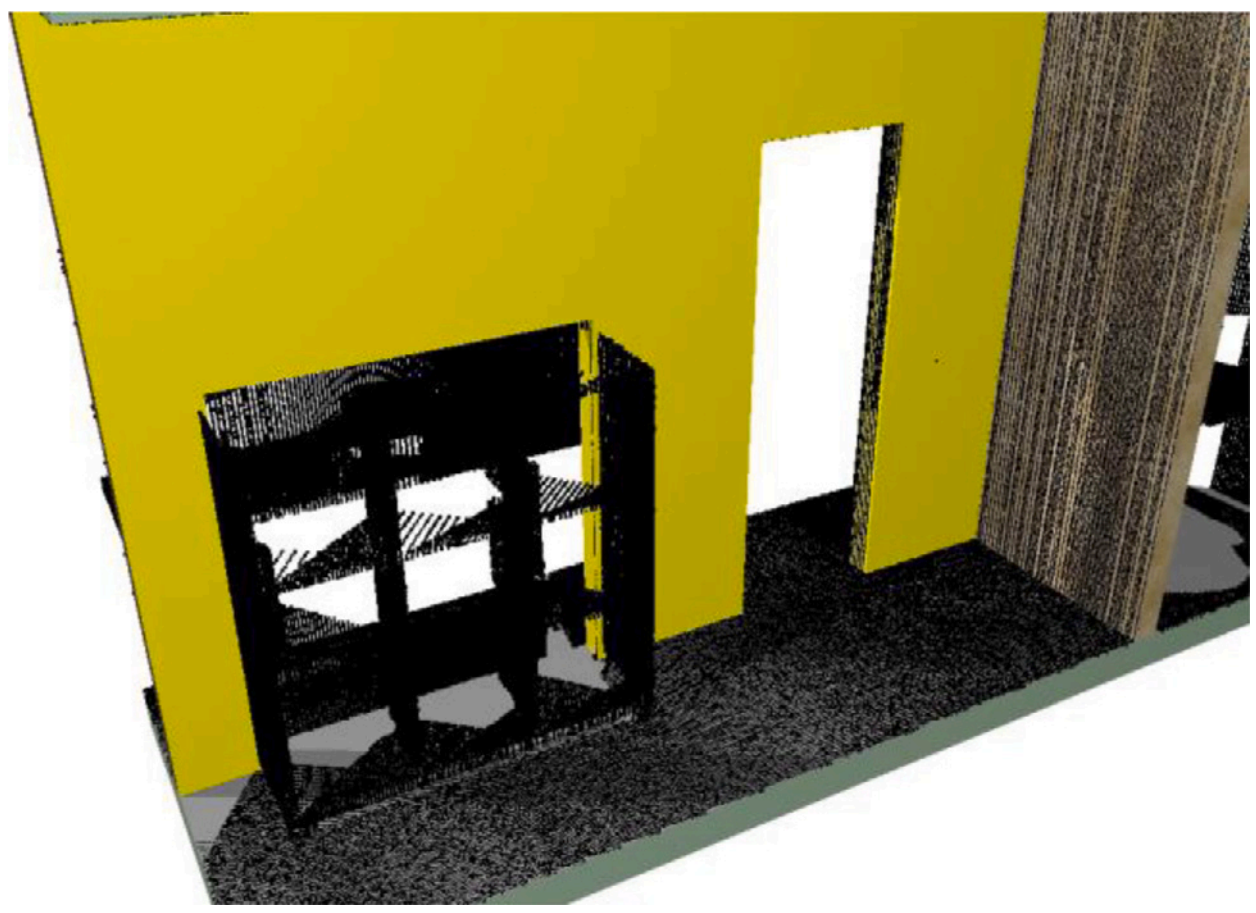

**Fig. 25.** Impact of obstruction on point cloud interpretation, causing false detection of a door opening in the reconstructed IFC model.

Third, the method assumes a uniform height level for each floor, which may lead to inaccuracies in buildings with more complex layouts. Features such as stepped floors, mezzanines, and varying ceiling heights within the same story are not currently accounted for in the segmentation process. As a result, the accuracy of the generated BIM models may be reduced in buildings with split-level designs or other non-standard height configurations.

Finally, the accuracy and reliability of the method are highly dependent on the quality of the input point cloud. If the scanned data is noisy, sparse, or contains excessive occlusions due to furniture or other obstructions (Fig. 25), the reconstruction process becomes significantly more challenging. In such cases, manual adjustments such as removal of incorrectly identified openings must be done in a BIM modeling software. To mitigate these challenges, high-quality point cloud acquisition is essential, including proper scanning techniques and preprocessing steps to filter out unwanted elements.

### *3.5. Final remarks and future developments*

Cloud2BIM is still evolving, with ongoing enhancements such as the integration of furniture recognition using deep-learning models [59]. Furthermore, we are developing a graphical user interface that will allow users to smoothly adjust and calibrate input parameters. This interface will feature live visualization, enabling immediate observation of how changes to calibrated parameters affect the model, thus ensuring precise and user-friendly software interaction.

Given the popularity of the RANSAC algorithm for segmenting planar elements from point clouds, one might question why it was not employed in our development. The decision was based on significant concerns regarding performance and the calibration of inputs. While RANSAC is adept at managing noise and outliers, its iterative and random sampling method can inadvertently produce spurious planes that do not correspond to actual planar surfaces [60], especially in complex indoor environments. This issue is compounded by the inherent uncertainty in the algorithm's hypothesis generation phase [61]. Although robust methods are designed to address these inaccuracies, they typically involve point-based operations that are inefficient for indoor settings.

Building on the capabilities of our software, we are exploring technologies to further enrich the BIM models with detailed features and materials not discernible through geometry alone. Inspired by Adán et al. [46], who utilized RGB data from point clouds to detect minor architectural elements like electrical sockets, we plan to adapt similar image analysis techniques. Although their approach currently lacks IFC export functionality, restricting its integration with BIM tools, it provides a promising avenue for adding realistic details such as color textures and material finishes to our models. This enhancement will not only increase the richness of the visual and functional aspects of the models but also supports comprehensive documentation and precise refurbishment planning.

Moreover, considering recent advancements in autonomous mobile scanning for 3D reconstruction, as proposed by Hu et al. [62], there is potential to extend Cloud2BIM or its components to support robot-assisted autonomous scanning workflows. This would enable instant point cloud segmentation and automated 3D model reconstruction directly at the point of data capture, significantly speeding up the BIM generation process and reducing the need for post-processing.

## 4. Conclusion

This paper introduced Cloud2BIM, a Python-based open-source automated scan-to-BIM pipeline that converts various point cloud formats into detailed 3D BIM models compliant with the IFC standard. Our approach is distinguished by a volumetric methodology that prioritizes capturing accurately the geometry of building elements over their mere spatial arrangement.

The segmentation and classification process within Cloud2BIM employs specialized algorithms tailored to each building element type, ensuring high accuracy and efficiency; Cloud2BIM is able to segment and classify:

- **Slabs:** The software combines density-based height segmentation, morphological operations, and contour detection to identify floor and ceiling slabs reliably. This method ensures accurate identification even in indoor-only environments without scanning facades or roofs.
- **Walls:** Wall detection is handled by a hybrid approach that combines the strengths of both room-based and wall-based segmentation. This strategy allows for precise transformation of point cloud data, accommodating different wall orientations and configurations without the need for exterior scans.
- **Openings:** Door and window openings are addressed through a histogram-based technique, enhanced with heuristic rules that evaluate the sill positions and the aspect ratio and dimensions of the openings. This approach ensures that each opening is accurately positioned and sized within the BIM model, taking into account architectural standards and construction practices.
- **Zones** are innovatively defined using actual wall surfaces rather than simple axes. This feature allows for, e.g., detailed definitions of the acoustic, thermal, and usage characteristics of individual rooms. Cloud2BIM creates representation of zones in IFC, opening the possibility for implementing non-graphical information into the BIM model.

Cloud2BIM has been extensively tested on various datasets and has demonstrated high accuracy and superior speed; for example, on the ETH Zurich synth3 dataset, Cloud2BIM achieved millimeter-level accuracy in aligning points to modeled surfaces and surpassed the fastest benchmark solutions by the factor of 7 while providing more functionalities.

As a community-driven platform, Cloud2BIM encourages continuous enhancement and collaborative development. It aims to integrate future technologies such as machine learning for furniture recognition and augmented reality for enriched contextual data. This integration aims to streamline the creation of detailed BIM documentation and make it more accessible to a wider audience, including small stakeholders and research institutions.

## CRediT authorship contribution statement

**Slávek Zbirovský:** Writing – original draft, Visualization, Software, Resources, Methodology, Investigation, Data curation. **Václav Nežerka:** Writing – review & editing, Supervision, Software, Project administration, Methodology, Investigation, Conceptualization.

## Declaration of generative AI and AI-assisted technologies in the writing process

During the preparation of this work, the authors used AI language model ChatGPT 4 from OpenAI L.L.C. (CA 94110 San Francisco, USA) in order to improve readability and language of the manuscript. After using these tools, the authors reviewed and edited the content as needed and take full responsibility for the content of the publication.

## Funding

This work was funded by the European Union under the project ROBOPROX (no. CZ.02.01.01/00/22_008/0004590), by the European Union's Horizon Europe Framework Programme (call HORIZON-CL4-2021-TWIN-TRANSITION-01-11) under grant agreement No. 10105 8580, project RECONMATIC (Automated solutions for sustainable and circular construction and demolition waste management), and the Czech Technical University in Prague, grant agreement No. SGS25/ 005/OHK1/1T/11.

## Declaration of competing interest

All the authors manuscript (such as honoraria; educational grants; participation in speakers bureaus; membership, employment, consultancies, stock ownership, or other equity interest; and expert testimony or patent-licensing arrangements), or non-financial interest (such as personal or professional relationships, affiliations, knowledge or beliefs) in the subject matter or materials discussed in this manuscript.

## Acknowledgments

We acknowledge the Visualization and MultiMedia Lab at the University of Zurich and Claudio Mura for acquiring the 3D point clouds. We also thank UZH and ETH Zurich for their support in scanning the rooms represented in these datasets. The synth3 dataset they provided was used for validation and testing of our algorithms. Special thanks go to Jan Zeman for consultations and editing and to Rudolf Urban and Zdeněk Svatý for providing us the Kladno station and Prague hotel datasets.

## Data availability

The data (point clouds and codes) are available under the link provided in the paper.